\documentclass[10pt,journal,compsoc]{IEEEtran}

\usepackage{xr}
\usepackage{amsmath}
\usepackage{amssymb}
\usepackage{pifont}

\usepackage{comment}
\usepackage{soul}
\usepackage{algorithmic}
\usepackage{hyperref}
\usepackage{booktabs}
\usepackage{threeparttable}
\usepackage{mathrsfs}
\usepackage{amsfonts}
\usepackage[ruled]{algorithm2e}

\usepackage{multirow}
\usepackage{graphicx}
\usepackage{color}
\usepackage{balance}
\usepackage{caption}
\usepackage{adjustbox}
\usepackage{soul}
\setstcolor{red}
\usepackage{array}
\usepackage{blindtext}
\usepackage{utfsym}

\usepackage{multirow}
\usepackage{tabularx}
\usepackage{xspace}
\usepackage{array}
\usepackage{caption}
\usepackage{subcaption}
\usepackage{soul}
\usepackage{enumitem}
\usepackage{wrapfig}
\usepackage{xcolor}
\usepackage{tikz}
\usetikzlibrary{tikzmark}

\usepackage{amsmath}
\usepackage{amssymb}
\usepackage{mathtools}
\usepackage{amsthm}

\newtheorem{theorem}{Theorem}
\newtheorem{definition}{Definition}
\newtheorem{proposition}{Proposition}
\newtheorem{lemma}{Lemma}

\newcommand{\definitionautorefname}{Definition}

\newcommand*{\shortautoref}[1]{%
  \begingroup
    \def\sectionautorefname{Sec.}%
    \def\subsectionautorefname{Sec.}%
    \def\figureautorefname{Fig.}%
    \def\tableautorefname{Tab.}%
    \def\equationautorefname{Eq.}%
    \def\subfigureautorefname{Fig.}%
    \def\definitionautorefname{Def.}%
    \autoref{#1}%
  \endgroup
}

\definecolor{uclablue}{rgb}{0.33, 0.41, 0.58}
\definecolor{azure2}{rgb}{0.0, 0.5, 1.0}
\definecolor{azure}{rgb}{0.19, 0.55, 0.91}
\definecolor{blazeorange}{rgb}{1.0, 0.4, 0.0}
\definecolor{forestgreen}{rgb}{0.13, 0.55, 0.13}
\definecolor{pink}{rgb}{1, 0, 0.5}
\definecolor{airforceblue}{rgb}{0.36, 0.54, 0.66}
\definecolor{tabhighlight}{HTML}{e5e5e5}
\definecolor{darkgrey}{rgb}{0.53,0.53,0.53}
\definecolor{mygrey}{rgb}{0.9,0.9,0.9}

\hypersetup{
    linkbordercolor=airforceblue,
    urlbordercolor=pink,
    citebordercolor=green,
}

\newcommand{\revision}{\textcolor{black}}
\newcommand{\revisiontwo}{\textcolor{black}}

\newcommand\SpaTemGNNs{\textsc{STGNNs}\xspace}
\newcommand\MPSTGNN{\textsc{MP-STGNNs}\xspace}
\newcommand\SpecTemGNNs{\textsc{SPTGNNs}\xspace}
\newcommand\SpecTemGNN{\textsc{SPTGNN}\xspace}
\newcommand\TGC{\textsc{TGGC}\xspace}
\newcommand\TGCfull{\textsc{TGGC}$^{\dagger}$\xspace}

\usepackage{tcolorbox}
\usepackage{colortbl}
\tcbuselibrary{skins}

\tcbset{tab2/.style={enhanced,fonttitle=\bfseries,
colback=white,colframe=gray,colbacktitle=white,
coltitle=black,center title}}

\tikzset{mycircled/.style={circle,draw,inner sep=0.05em,line width=0.04em, scale=0.8}}

\usepackage{makecell}
\ifCLASSOPTIONcompsoc
  \usepackage[nocompress]{cite}
\else
  \usepackage{cite}
\fi

\ifCLASSINFOpdf

\else

\fi

\begin{document}

\title{Towards Expressive Spectral-Temporal Graph Neural Networks for Time Series Forecasting}

\author{
Ming Jin, 
Guangsi Shi, 
Yuan-Fang Li,
Bo Xiong,
Tian Zhou,
Flora D. Salim,
Liang Zhao,~\IEEEmembership{Senior Member,~IEEE},
Lingfei Wu,
Qingsong Wen,~\IEEEmembership{Senior Member,~IEEE},
Shirui Pan,~\IEEEmembership{Senior Member,~IEEE}

\IEEEcompsocitemizethanks{
\IEEEcompsocthanksitem Ming Jin and Shirui Pan are with the School of Information and Communication Technology, Griffith University, Gold Coast, Australia. E-mail: mingjinedu@gmail.com,   s.pan@griffith.edu.au;
\IEEEcompsocthanksitem Guangsi Shi and Yuan-Fang Li are with the Department of Data Science and AI, Monash University, Melbourne, Australia. E-mail: \{guangsi.shi, yuanfang.li\}@monash.edu;
\IEEEcompsocthanksitem Bo Xiong is with International Max Plank Research School for Intelligent Systems and the University of Stuttgart, Stuttgart, Germany. E-mail: bo.xiong@ki.uni-stuttgart.de;
\IEEEcompsocthanksitem Tian Zhou is with Alibaba Group, Hangzhou, China. E-mail: tian.zt@alibaba-inc.com;
\IEEEcompsocthanksitem Flora D. Salim is with the School of Computer Science and Engineering, University of New South Wales, Sydney, Australia. E-mail: flora.salim@unsw.edu.au;
\IEEEcompsocthanksitem Liang Zhao is with the Department of Computer Science at Emory University, Atlanta, USA. E-mail: liang.zhao@emory.edu.
\IEEEcompsocthanksitem Lingfei Wu is with Anytime.AI, New York, USA. E-mail: teddy.lfwu@gmail.com. 
\IEEEcompsocthanksitem Qingsong Wen is with Squirrel Ai Learning, Bellevue, US. E-mail: qingsongedu@gmail.com.
\IEEEcompsocthanksitem Ming Jin and Guangsi Shi contributed equally to this work.
}
\thanks{Corresponding Authors: Qingsong Wen, Shirui Pan.}
}

\markboth{Journal of \LaTeX\ Class Files,~Vol.~14, No.~8, August~2021}%
{Shell \MakeLowercase{\textit{et al.}}: Bare Demo of IEEEtran.cls for Computer Society Journals}

\IEEEtitleabstractindextext{%
\begin{abstract}
Time series forecasting has remained a focal point due to its vital applications in sectors such as energy management and transportation planning. Spectral-temporal graph neural network is a promising abstraction underlying most time series forecasting models that are based on graph neural networks (GNNs). However, more is needed to know about the underpinnings of this branch of methods. In this paper, we establish a theoretical framework that unravels the expressive power of spectral-temporal GNNs. Our results show that linear spectral-temporal GNNs are universal under mild assumptions, and their expressive power is bounded by our extended first-order Weisfeiler-Leman algorithm on discrete-time dynamic graphs. To make our findings useful in practice on valid instantiations, we discuss related constraints in detail and outline a theoretical blueprint for designing spatial and temporal modules in spectral domains. Building on these insights and to demonstrate how powerful spectral-temporal GNNs are based on our framework, we propose a simple instantiation named \textit{Temporal Graph Gegenbauer Convolution} (TGGC), which significantly outperforms most existing models with only linear components and shows better model efficiency. Our findings pave the way for devising a broader array of provably expressive GNN-based models for time series.
\end{abstract}

}

\maketitle

\IEEEdisplaynontitleabstractindextext

\IEEEpeerreviewmaketitle

\section{Introduction}
\begin{figure}[t]
    \centering
    \includegraphics[width=0.45\textwidth]{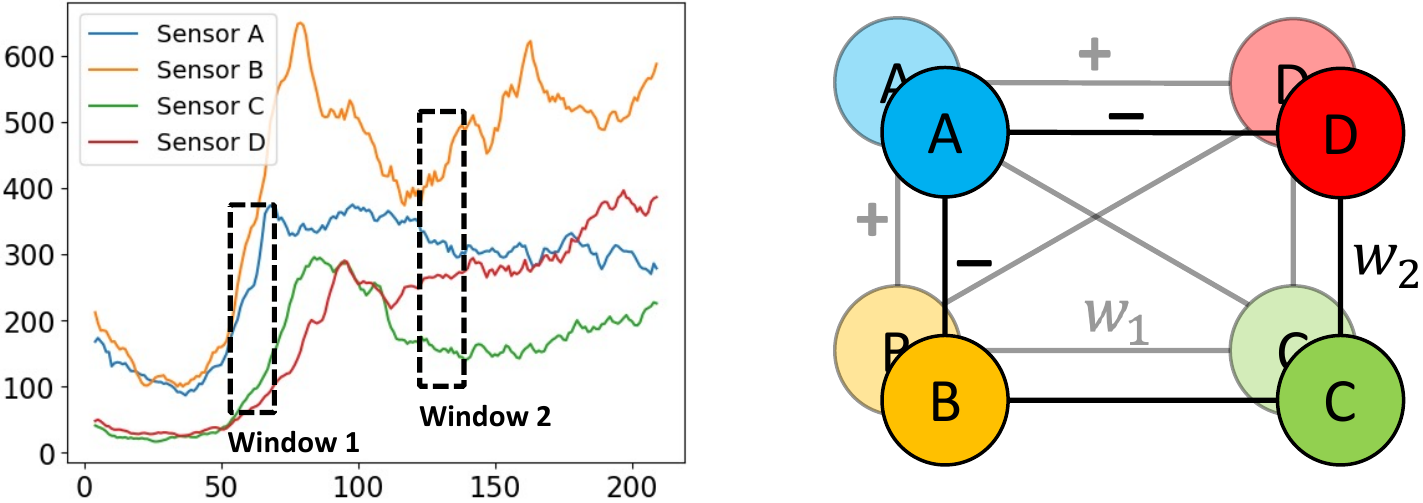}
    \caption{
    Differently signed spatial relations between time series in a real-world traffic (PeMS07) dataset. \textbf{Left}: Visualization of four randomly selected traffic sensor readings. \textbf{Right}: Spatial relations between time series may be different in two windows (e.g., A and B are positively and negatively correlated in windows 1 and 2, respectively), where the (weighted) edges between the sensors normally indicate their correlation strengths but not reflect the signs.
    Unlike most spatio-temporal GNNs that aggregate the neighborhood information without considering correlation signs, spectral-temporal GNNs go beyond low-pass filtering by learning to aggregate or differentiate such information.
    }
    \label{fig: introduction signed graph}
\end{figure}

Graph neural networks (GNNs) have achieved considerable success in static graph representation learning for many tasks \cite{hamilton2020graph}. Many existing studies, such as STGCN \cite{yu2018spatio} and Graph WaveNet \cite{wu2019graph}, have successfully extended GNNs to time series forecasting. Among these methods, either first-order approximation of ChebyConv \cite{defferrard2016convolutional} or graph diffusion \cite{klicpera2019diffusion} is typically used to model time series relations, \revisiontwo{where a strong assumption of local homophiles -- a property where connected nodes are assumed to have similar features or states -- is made \cite{li2021beyond}. This assumption is inherent in these methods where the graph convolution is approximated by aggregating information from neighboring nodes.} Thus, they are only capable of modeling \textbf{\textit{positive correlations}} between time series that exhibit strong similarities, and we denote this branch of methods as message-passing-based spatio-temporal GNNs (\MPSTGNN). Nevertheless, how to model real-world multivariate time series with complex spatial dependencies that evolve remains an open question. This complexity is depicted in \shortautoref{fig: introduction signed graph} within a widely adopted real-world traffic dataset, in which \textbf{\textit{differently signed relations}} between time series are evident. For example, in the first window ($w_1$) of traffic volume, the readings from sensor A and B are positively correlated, but this is not the case in the second window ($w_2$) of data.

Existing \MPSTGNN often struggle with capturing complex, dynamic relationships in time series data, and their limitations stem from a lack of expressiveness, \revisiontwo{which we define as the ability of a model to accurately represent a wide range of temporal and spatial dependencies within the data. The expressive power of a spatio-temporal GNN is crucial for effective time series forecasting because it determines the model’s capacity to capture intricate patterns and relationships that are inherent in real-world time series.} \revisiontwo{Spectral-temporal GNNs (\SpecTemGNNs), as an advanced abstraction of \MPSTGNN, shed light on modeling differently signed time series correlations by approximating graph convolutions with a broad range of graph spectral filters beyond low-pass filtering \cite{he2021bernnet, derr2018signed}.} This is evidenced by a recent work \cite{cao2020spectral} in multivariate time series forecasting. Although it demonstrated competitive performance, the theoretical foundations of \SpecTemGNNs remain under-researched, which hinders the understanding of \SpecTemGNNs and the development of follow-up research within this model family. Accordingly, in this research, we identify several unresolved fundamental questions:

\begin{tcolorbox}[notitle, rounded corners, colframe=darkgrey, colback=white, boxrule=2pt, boxsep=0pt, left=0.15cm, right=0.15cm, enhanced, shadow={2.5pt}{-2.5pt}{0pt}{opacity=5,mygrey},toprule=2pt, before skip=0.65em, after skip=0.75em, width=0.98\linewidth
  ]
  {
    \centering 
  {
    \fontsize{8pt}{13.2pt}\selectfont 
    \vspace{2mm}
    \begin{description}[nosep]
    \item[Q1.] What is the general formulation of \SpecTemGNNs? 
    \item[Q2.] How expressive are \SpecTemGNNs in modeling time series data?
    \item[Q3.] When does this model family fail to generalize well?
    \item[Q4.] How to design provably expressive \SpecTemGNNs for effective time series forecasting?
    \end{description} 
    \vspace{2mm}
  }
  }
\end{tcolorbox}

In this work, we establish a series of theoretical results summarized in \shortautoref{fig: schematic diagram and contributions} to answer these questions. We begin by formulating a general framework of \SpecTemGNNs (\textcolor{black}{\textbf{Q1}}; \shortautoref{subsec: general formulation of spectral-temporal GNNs}), and then prove its universality of linear models (i.e., linear \SpecTemGNNs are powerful enough to represent arbitrary time series) under mild assumptions through the lens of discrete-time dynamic graphs (DTDGs) \cite{jin2022multivariate} \nocite{luo2023graph, jin2022neural} and spectral GNNs \cite{wang22jacobi}. We further discuss related constraints from various aspects (\textcolor{black}{\textbf{Q3}}; \shortautoref{subsec:expressive power analysis}) to make our theorem useful in practice on \textbf{\textit{any}} valid instantiations. After this, we extend the classic color-refinement algorithm~\cite{kiefer2020power} on DTDGs and prove that the expressive power of \SpecTemGNNs is theoretically bounded by the proposed temporal 1-WL test (\textcolor{black}{\textbf{Q2}}; \shortautoref{subsec:expressive power analysis}). To answer the last question (\textcolor{black}{\textbf{Q4}}; \shortautoref{subsec:design of spectral filters}), we prove that under mild assumptions, linear \SpecTemGNNs are sufficient to produce expressive time series representations with orthogonal function bases and individual spectral filters in their graph and temporal frequency-domain models. Our results, for the first time, unravel the capabilities of \SpecTemGNNs and outline a blueprint for designing powerful GNN-based time series models.

Drawing from and to validate these theoretical insights, we present a straightforward, yet effective and novel \SpecTemGNN instantiation, named \TGC (short for \textit{Temporal Graph Gegenbauer Convolution}), that well generalizes the related work in time series forecasting, such as STGCN~\cite{yu2018spatio} and StemGNN~\cite{cao2020spectral}. Though our primary goal is not to achieve state-of-the-art performance, our method, remarkably, is very efficient and significantly outperforms numerous existing models on several time series benchmarks and forecasting settings \textbf{\textit{with minimal designs}}. The proposed \TGC building block encompasses only simple linear spatial and temporal frequency-domain models. Comprehensive experiments on synthetic and real-world datasets demonstrate that: \textbf{(1)} our approach excels at learning time series relations of different signs compared to \MPSTGNN; \textbf{(2)} our design principles, e.g., orthogonal bases and individual filtering, are crucial for \SpecTemGNNs to perform well; \textbf{(3)} our instantiation (\TGC) can be readily augmented with nonlinearities and other common model choices. Finally, and more importantly, our findings pave the way for devising a broader array of provably expressive \SpecTemGNNs and thus shed light on subsequent research. Our main contributions in this work are summarized as follows.

\begin{itemize}
    \item To our knowledge, we are the first to theoretically ground spatio-temporal graph learning in the context of time series forecasting using the frequency analysis. This proves the expressiveness of this family of methods for time series modeling and helps identify situations in which they may fall short.
    \item We establish a general framework for spectral-temporal graph neural networks (\SpecTemGNNs) and answer the question of how to design provably expressive \SpecTemGNNs for effective time series forecasting. This is exemplified by our simple, yet effective and novel implementation, named \textit{Temporal Graph Gegenbauer Convolution} (\TGC), which contains only linear components and operates entirely in the frequency domain.
    \item We demonstrate that \TGC can outperform the most representative and related models, and its advanced nonlinear variant (\TGCfull) consistently surpasses other competitive counterparts. Our comprehensive evaluation also explicitly demonstrates that our approach excels at learning time series relations of different signs.
\end{itemize}

\begin{figure}[t]
\centering
\input{tables/contributions}
\vspace{-2mm}
\caption{An overview of the theoretical results in this work.}
\label{fig: schematic diagram and contributions}
\vspace{-3mm}
\end{figure}

The structure of this paper is outlined as follows: \shortautoref{sec:related_work} provides an overview of the related work, examining various facets in detail. \shortautoref{sec:preliminary} delineates the notations and introduces essential background knowledge. \shortautoref{sec:sptgnns} presents a theoretical framework to deepen the understanding of \SpecTemGNNs, addressing the research questions set forth. \shortautoref{sec:instant-tgc} showcases our theoretical insights through a novel and straightforward implementation known as \TGC, including its advanced variant \TGCfull. Our research findings are evaluated and shared in \shortautoref{sec:evaluation}, and a concise conclusion encapsulating our discoveries can be found in \shortautoref{sec:conclusion}.
\section{Related Work}\label{sec:related_work}
In this section, we provide a concise overview of pertinent literature, encompassing contemporary time series forecasting models, spatio-temporal graph neural networks, and recent advancements in spectral graph neural networks.

\subsection{Deep Time Series Forecasting}
Time series forecasting has been extensively researched over time. Traditional approaches primarily focus on statistical models, such as vector autoregressive (VAR)~\cite{lutkepohl2013vector} and autoregressive integrated moving average (ARIMA)~\cite{box2015time}. Deep learning-based approaches, on the other hand, have achieved great success in recent years. For example, recurrent neural network (RNN) and its variants, e.g., FC-LSTM~\cite{shi2015convolutional}, are capable to well model univariate time series. TCN~\cite{bai2018empirical} improves these methods by modeling multivariate time series as a unified entity and considering the dependencies between different variables. Follow-up research, such as LSTNet~\cite{lai2018modeling} and DeepState~\cite{rangapuram2018deep}, proposes more complex models to handle interlaced temporal and spatial clues by marrying sequential models with convolution networks or state space models. Recently, Transformer~\cite{vaswani2017attention}-based approaches have made great leaps, especially in long-term forecasting~\cite{wen2022transformers}. For these methods, an encoder-decoder architecture is normally applied with improved self- and cross-attention, e.g., logsparse attention~\cite{li2019enhancing}, locality-sensitive hashing~\cite{kitaev2019reformer}, and probability sparse attention~\cite{zhou2021informer}. As time series can be viewed as a signal of mixed seasonalities, Zhou \textit{et al.} further propose FEDformer~\cite{zhou22fedformer} and a follow-up work FiLM~\cite{zhou2022film} to rethink how spectral analysis benefits time series forecasting. Nevertheless, these methods do not explicitly model inter-time series relationships (i.e., spatial dependencies).

\subsection{Spatio-Temporal Graph Neural Networks}
A line of research explores capturing time series relations using GNNs. For instance, DCRNN~\cite{li2018diffusion} combines recurrent units with graph diffusion~\cite{klicpera2019diffusion} to simultaneously capture temporal and spatial dependencies, while Graph WaveNet~\cite{wu2019graph} interleaves TCN~\cite{bai2018empirical} and graph diffusion layers. Subsequent studies, such as STSGCN~\cite{sofianos2021space}, STFGNN~\cite{li2021spatial}, STGODE~\cite{fang2021spatial} and G-SWaN~\cite{prabowo2023because}, adopt similar principles but other ingenious designs to better characterize underlying spatio-temporal clues. However, these methods struggle to model differently signed time series relations since their graph convolutions operate under the umbrella of message passing, which serves as low-pass filtering assuming local homophiles. Some \SpaTemGNNs, such as ASTGCN~\cite{guo2019attention} and LSGCN~\cite{huang2020lsgcn}, directly employ ChebConv~\cite{defferrard2016convolutional} for capturing time series dependencies. However, their graph convolutions using the Chebyshev basis, along with the intuitive temporal models they utilize, result in sub-optimal solutions. Consequently, the expressiveness of most \SpaTemGNNs remains limited. Spectral-temporal graph neural networks (\SpecTemGNNs), on the other hand, first make it possible to fill the gap by (properly) approximating both graph and temporal convolutions with a broad range of filters in spectral domains, allowing for more accurate pattern extraction and modeling. A representative work in this category is StemGNN~\cite{cao2020spectral}. However, it faces two fundamental limitations: \textbf{(1)} its direct application of ChebConv is sub-optimal, and \textbf{(2)} although its temporal FDMs employ orthogonal space projections, they fail to make proper multidimensional and multivariate predictions as discussed.

\subsection{Spectral Graph Neural Networks}
Spectral graph neural networks are grounded in spectral graph signal filtering, where graph convolutions are approximated by truncated polynomials with finite degrees. These graph spectral filters can be either trainable or not. Examples with predefined spectral filters include APPNP~\cite{gasteiger2018predict} and GraphHeat~\cite{xu2019graph}, as illustrated in~\cite{he2021bernnet}. Another branch of work employs different polynomials with trainable coefficients (i.e., filter weights) to approximate effective graph convolutions. For instance, ChebConv~\cite{defferrard2016convolutional} utilizes Chebyshev polynomials, inspiring the development of many popular spatial GNNs with simplifications~\cite{kipf2016semi, luo2023npfkgc}. BernNet~\cite{he2021bernnet} employs Bernstein polynomials but can only express positive filter functions due to regularization constraints. Another example is \cite{bianchi2021graph} that implements the graph convolution with auto-regressive moving average (ARMA) filters. Recently, JacobiConv~\cite{wang22jacobi} demonstrated that graph spectral filtering with Jacobi polynomial approximation is highly effective on a wide range of graphs under mild constraints.  Although spectral GNNs pave the way for \SpecTemGNNs, they primarily focus on modeling static graph-structured data without the knowledge telling how to effectively convolute on dynamic graphs for modeling time series data.
\section{Preliminary}\label{sec:preliminary}
In time series forecasting, given a series of historical observations $\mathbf{X} \in \mathbb{R}^{N \times T \times D}$ encompassing $N$ different $D$-dimensional variables across $T$ time steps, we aim to learn a function $f(\cdot): \mathbf{X} \mapsto  \Hat{\mathbf{Y}}$, where the errors between the forecasting results $\Hat{\mathbf{Y}} \in \mathbb{R}^{N \times H \times D}$ and ground-truth $\mathbf{Y}$ are minimized with the following mean squared loss: $\frac{1}{H} \sum_{t=1}^{H} || \Hat{\mathbf{Y}}_t - \mathbf{Y}_t ||^2_F$. $H$ denotes the forecasting horizon.
In this work, we learn an adjacency matrix $\mathbf{A} \in \mathbb{R}^{N \times N}$ from the input window $\mathbf{X}$ to describe the connection strength between $N$ variables, as in \cite{cao2020spectral}. We use $\mathbf{X}_t := \mathbf{X}_{:,t,:} \in \mathbb{R}^{N \times D}$ to denote the observations at a specific time $t$, and $\mathbf{X}_n := \mathbf{X}_{n,:,:} \in \mathbb{R}^{T \times D}$ as a time series of a specific variable $n$ with $T$ time steps and $D$ feature dimensions.  \\

\noindent\textbf{Graph Spectral Filtering.}
For simplicity and modeling the multivariate time series from the graph perspective, we let $\mathcal{G}_t=(\mathbf{A}, \mathbf{X}_t)$ denote an \textbf{\textit{undirected}} graph snapshot at a specific time $t$ with the node features $\mathbf{X}_t$. In a graph snapshot, $\mathbf{A}$ and its degree matrix $\mathbf{D} \in \mathbb{R}^{N \times N}$ s.t. $\mathbf{D}_{i,i} = \sum_{j=1}^{N} \mathbf{A}_{i,j}$; thus, its normalized graph Laplacian matrix $\hat{\mathbf{L}} = \mathbf{D}^{-\frac{1}{2}}(\mathbf{D} - \mathbf{A})\mathbf{D}^{-\frac{1}{2}}$ is symmetric and can be proven to be positive semi-definite. We let its eigendecomposition of $\hat{\mathbf{L}}$ to be $\hat{\mathbf{L}} = \mathbf{U}\mathbf{\Lambda}\mathbf{U}^\top$, where $\mathbf{\Lambda}$ and $\mathbf{U}$ are matrices of eigenvalues and eigenvectors. Below, we define the graph convolution to filter the input signals in the spectral domain w.r.t.\ the node connectivity.

\begin{definition}[Graph Convolution]\label{def:graph_convolution}
Assume there is a filter function of eigenvalues $g(\cdot): [0, 2] \mapsto \mathbb{R}$, we define the graph convolution on $\mathcal{G}_t$ as filtering the input signal $\mathbf{X}_t$ with the spectral filter:
\begin{equation}
    g(\mathbf{\Lambda}) \star \mathbf{X}_t := \mathbf{U}g(\mathbf{\Lambda})\mathbf{U}^\top\mathbf{X}_t.
\label{eq: spectral graph convolution}
\end{equation}
\end{definition}

\revision{The $\star$ operator is rooted in the graph Fourier transform, where the frequency components of input signal $\mathbf{U}^\top\mathbf{X}_t$ are transformed with $g(\mathbf{\Lambda})$ before being projected back using $\mathbf{U}$. However, directly computing \shortautoref{eq: spectral graph convolution} can be computationally expensive due to the eigendecomposition involved. To address this,} we approximate the \textbf{\textit{learnable}} filter function $g_{\theta}(\mathbf{\Lambda})$ with a truncated $K$-degree polynomial expansion~\cite{zheng2023node}: 
\begin{equation}
    g_{\theta}(\lambda) := \sum_{k=0}^{K} \mathbf{\Theta}_{k,:} P_k(\lambda).
\end{equation}

Thus, the graph spectral filtering takes the form:
\begin{equation}
    \mathbf{U}g_{\theta}(\mathbf{\Lambda})\mathbf{U}^\top\mathbf{X}_t \!:= \! \sum_{k=0}^{K} \mathbf{\Theta}_{k,:} \mathbf{U} P_k(\mathbf{\Lambda}) \mathbf{U}^\top \mathbf{X}_t\! = \!\!\sum_{k=0}^{K} \mathbf{\Theta}_{k,:} P_k(\hat{\mathbf{L}}) \mathbf{X}_t.
\end{equation}

If we let $g_{\theta}(\hat{\mathbf{L}}) = \sum_{k=0}^{K} \mathbf{\Theta}_{k,:} P_k(\hat{\mathbf{L}})$, then the graph convolution is redefined as: $g(\mathbf{\Lambda}) \star \mathbf{X}_t := g_{\theta}(\hat{\mathbf{L}}) \mathbf{X}_t$. \\

\noindent\textbf{Orthogonal Time Series Representations.}
Time series can be analyzed in both time and spectral domains. In this work, we focus on modeling time series using sparse orthogonal representations, \revision{which are effective in helping disentangle and identify underlying patterns such as periodicities and in reducing modeling complexity}. Specifically, for an input signal of the $n^{\text{th}}$ variable, we represent $\mathbf{X}_{n}$ by a set of orthogonal components $\Tilde{\mathbf{X}}_n$. As per \autoref{lemma: time series approximation}, numerous orthogonal projections can serve as space projections for our objectives, e.g., \cite{wang2018packing}. In this study, we employ Fourier transformations by default in our proposed method (\shortautoref{sec:instant-tgc}).

\begin{definition}
Discrete Fourier transformation (DFT) on time series takes measurements at discrete intervals, and transforms observations into frequency-dependent amplitudes:
\begin{equation}
\small
    \Tilde{\mathbf{X}}_n(k) = \sum_{t=0}^{T-1} \mathbf{X}_n(t) e^{-2 \pi ikt / T},\ k = \{0, 1, \cdots, T-1\}.
\label{eq: discrete fourier transform}
\end{equation}
Inverse transformation (IDFT) maps a signal from the frequency domain back to the time domain:
\begin{equation}
\small
    \mathbf{X}_n(t) = \frac{1}{T} \sum_{k=0}^{T-1} \Tilde{\mathbf{X}}_n(k) e^{2 \pi ikt / T},\ t = \{0, 1, \cdots, T-1\}.
    \label{eq: inverse discrete fourier transform}
\end{equation}
\label{definition: discrete fourier transformation}
\end{definition}
\section{A Theoretical Framework of Spectral-Temporal Graph Neural Networks}\label{sec:sptgnns}
We address \textbf{\textit{all}} research questions in this section. We first introduce the general formulation of \SpecTemGNNs and then unravel the expressive power of this family of methods within an established theoretical framework. On this basis, we further shed light on the design of powerful \SpecTemGNNs with relevant theoretical justifications and proofs.

\subsection{General Formulation}\label{subsec: general formulation of spectral-temporal GNNs}

\begin{figure}[t]
    \centering
    \includegraphics[width=0.48\textwidth]{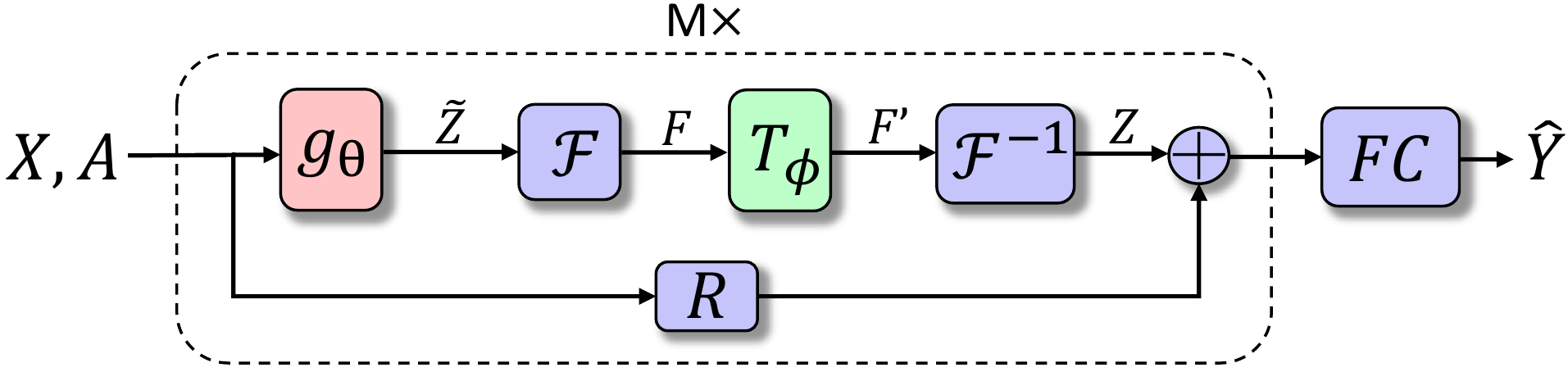}
    \caption{The general formulation of \SpecTemGNNs with $M$ building blocks to predict future values $\Hat{\mathbf{Y}}$ based on historical observations $\mathbf{X}$. We denote $g_\theta(\cdot)$ and $\mathcal{T}_\phi(\cdot)$ as graph and temporal spectral filters. $\mathcal{F}(\cdot)$ and $\mathcal{F}^{-1}(\cdot)$ are forward and inverse space projections.}
    \label{fig: stgnn framework}
\end{figure}

\noindent\textbf{Overall Architecture.} We illustrate the general formulation of \SpecTemGNNs in \shortautoref{fig: stgnn framework}, where we stack $M$ building blocks to capture spatial and temporal dependencies in spectral domains. Without loss of generality, we formulate this framework with the minimum redundancy and a straightforward optimization objective, where common add-ons in prior arts, e.g., spectral attention \cite{zhou22fedformer}, can be easily incorporated. 
To achieve time series connectivity without prior knowledge, we directly use the latent correlation layer from~\cite{cao2020spectral}, as this is not our primary focus. \\

\noindent\textbf{Building Block.} From the graph perspective and given the adjacency matrix $\mathbf{A}$, we can view the input signal $\mathbf{X}$ as a particular DTDG with a sequence of regularly-sampled static graph snapshots $\{\mathcal{G}_t\}_{t=0}^{T-1}$, where node features evolve but with fixed graph topology. In a \SpecTemGNN block, we filter $\mathbf{X}$ from the spatial and temporal perspectives in spectral domains with \textbf{\textit{graph spectral filters}} (GSFs) and \textbf{\textit{temporal spectral filters}} (TSFs). Formally, considering a single dimensional input $\mathbf{X}_d := \mathbf{X}_{:,:,d} \in \mathbb{R}^{N \times T}$, we define a \SpecTemGNN block as follows without the residual connection:
\begin{equation}
    \mathbf{Z}_d = \mathcal{T}_{\phi_d}\Big(g_{\theta_{d}}(\hat{\mathbf{L}})\mathbf{X}_d\Big) = \mathcal{T}_{\phi_d}\Big(\sum_{k=0}^{K} \mathbf{\Theta}_{k,d} P_k(\hat{\mathbf{L}}) \mathbf{X}_d\Big).
\label{eq: general formula}
\end{equation}
This formulation is straightforward. $\mathcal{T}_{\phi}(\cdot)$ represents TSFs, which work in conjunction with space projectors (detailed in \shortautoref{subsec:design of spectral filters}) to model temporal dependencies between node embeddings across snapshots. The internal expansion corresponds to GSFs' operation, which embeds node features in each snapshot $\mathcal{G}_t$ by learning variable relations.
The above process can be understood in different ways. The polynomial bases and coefficients generate distinct GSFs, allowing the internal $K$-degree expansion in \shortautoref{eq: general formula} to be seen as a combination of different \textbf{\textit{dynamic graph profiles}} at varying hops in the graph domain. Each profile filters out a specific frequency signal. Temporal dependencies are then modeled for each profile before aggregation: $\mathbf{Z}_d = \sum_{k=0}^{K} \mathcal{T}_{\phi_d}\Big(\mathbf{\Theta}_{k,d} P_k(\hat{\mathbf{L}}) \mathbf{X}_d\Big)$, which is equivalent to our formulation. Alternatively, dynamic graph profiles can be formed directly in the spectral domain, resulting in the formulation in StemGNN \cite{cao2020spectral} with increased model complexity.

\begin{figure*}[t]
     \centering
     \begin{subfigure}[b]{0.48\linewidth}
         \centering
         \includegraphics[width=\linewidth]{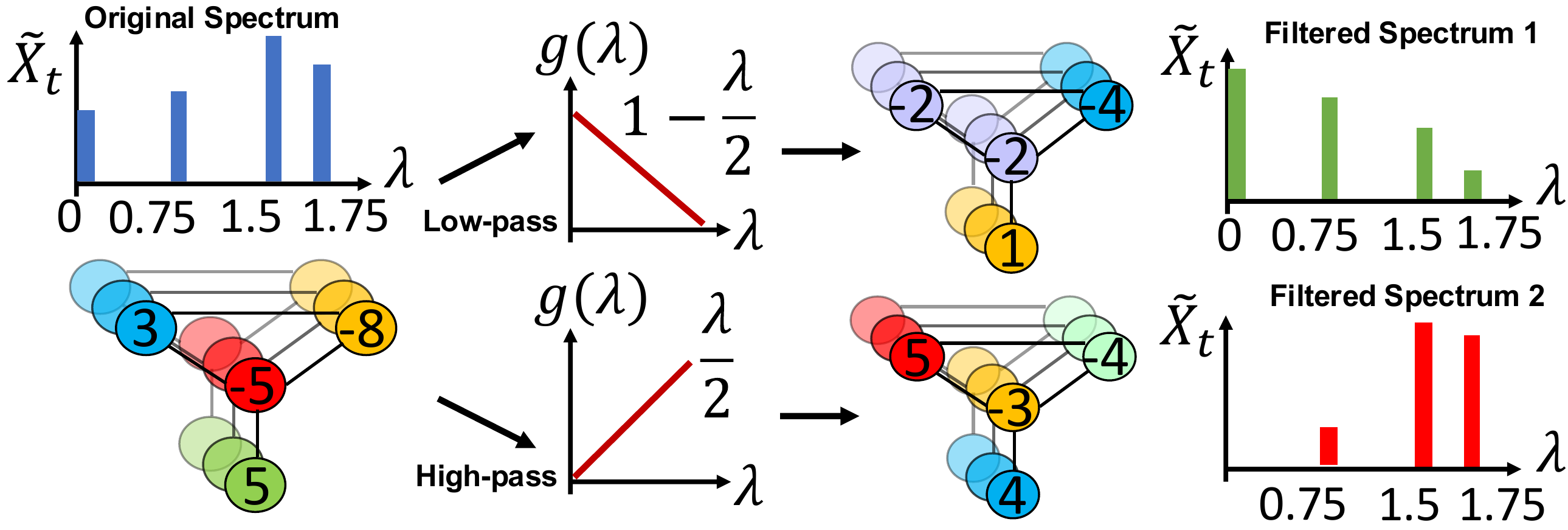}
     \end{subfigure}
     \hfill
     \begin{subfigure}[b]{0.48\linewidth}
         \centering
         \includegraphics[width=\linewidth]{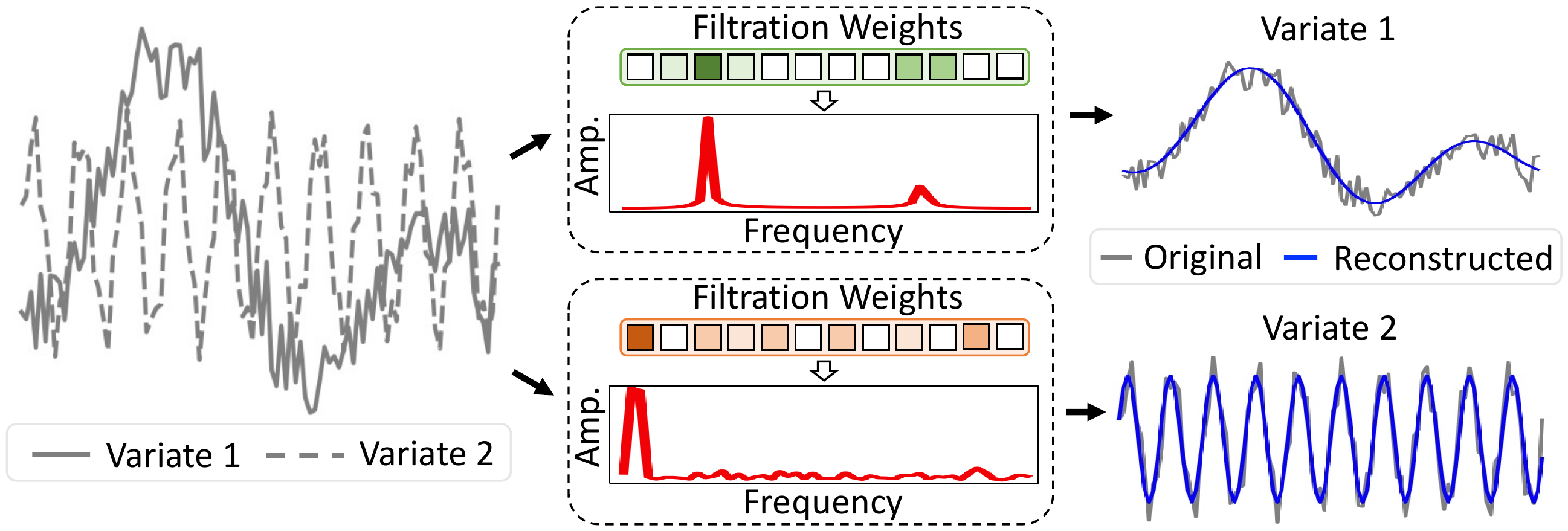}
     \end{subfigure}
     \caption{Multidimensional and multivariate predictions. \revision{\textbf{Left:} Multidimensional predictions within a snapshot require individual filtration for each output dimension to preserve different information.} \textbf{Right:} Individual filter is needed to model each single-dimensional time series.}
     \label{fig: multidimensional and multivariate predictions}
\end{figure*}

\subsection{Expressive Power of Spectral-Temporal GNNs}\label{subsec:expressive power analysis}
In this subsection, we develop a theoretical framework bridging spectral and dynamic GNNs, elucidating the expressive power of \SpecTemGNNs for modeling multivariate time series. All proofs are in \textbf{\textit{Appendix A}}. \\

\noindent\textbf{Linear GNNs.} \revision{For a linear GNN on $\mathbf{X} \in \mathbb{R}^{N \times D}$, we define it using a trainable weight matrix $\mathbf{W} \in \mathbb{R}^{D \times d}$ and parameterized spectral filters $g_{\theta}(\Hat{\mathbf{L}})$ as $\Tilde{\mathbf{Z}} = g_{\theta}(\Hat{\mathbf{L}}) \mathbf{X} \mathbf{W} \in \mathbb{R}^{N \times d}$.} This simple linear GNN can express any polynomial filter functions, denoted as polynomial-filter-most-expressive (PFME), under mild assumptions~\cite{wang22jacobi}. Its expressiveness establishes a lower bound for spectral GNNs. \\

To examine the expressive power of \SpecTemGNNs, we generalize spectral GNNs to model dynamic graphs. For simplicity, we initially consider linear GNNs with a single linear mapping $f_{\phi}(\cdot)$. To align with our formulation, $\mathcal{T}_\phi(\cdot)$ should also be linear functions; thus, \shortautoref{eq: general formula} can be interpreted as a linear GNN extension, where $f_{\phi}(\cdot)$ depends on historical observations instead of a single graph snapshot. Accordingly, linear \SpecTemGNNs establish a lower bound for the expressive power of \SpecTemGNNs.

\begin{proposition}
A \SpecTemGNN can differentiate any pair of nodes at an arbitrary valid time that linear \SpecTemGNNs can if $\mathcal{T}_\phi(\cdot)$ can express any linear time-variant functions.
\label{prop: linear stgnns form lower bound}
\end{proposition}

\revision{Despite their simplicity, linear \SpecTemGNNs retain the essential operations of spectral filtering, specifically the use of GSFs and TSFs that are integral to SPTGNNs, regardless of the complexity of the model.} We begin by defining the \textbf{\textit{universal approximation theorem}} for linear \SpecTemGNNs.

\begin{theorem} \label{theorem: linear stgnn} 
A linear \SpecTemGNN can produce arbitrary dimensional time series representations at any valid time iff: \textbf{(1)} $\Hat{\mathbf{L}}$ has no repeated eigenvalues; \textbf{(2)} $\mathbf{X}$ encompasses all frequency components with respect to the graph spectrum; \textbf{(3)} $\mathcal{T}_\phi(\cdot)$ can express any single-dimensional univariate time series.
\end{theorem}

Next, we separately explore these conditions and relate them to practical spectral-temporal GNNs. \\

\noindent\textbf{Multidimensional and Multivariate Prediction.} In a graph snapshot, each dimension may exhibit different properties, necessitating distinct filters for processing \cite{wang22jacobi}. An example with two-dimensional predictions is provided in \shortautoref{fig: multidimensional and multivariate predictions}. A simple solution involves using multidimensional polynomial coefficients, as explicitly shown in \shortautoref{eq: general formula}. A concrete example is presented in \cite{he2021bernnet}, where given a series of Bernstein bases, different polynomial coefficients result in various Bernstein approximations, corresponding to distinct GSFs. Similarly, when modeling temporal clues between graph snapshots, either dimension in any variable constitutes a unique time series requiring specific filtering. An example of reconstructing a two-variate time series of one dimension is in \shortautoref{fig: multidimensional and multivariate predictions}. In practice, we use multidimensional masking and weight matrices for each variable, forming a set of different TSFs (\shortautoref{subsec:design of spectral filters}). \\

\noindent\textbf{Frequency Components and Eigenvalues.} \revisiontwo{GSFs can only scale existing frequency components of specific eigenvalues. Frequency components sharing the same eigenvalue will be transformed by the same filter. For example, in a graph with repeated eigenvalues due to its topology (e.g., a highly symmetric graph with three- or higher-order automorphisms~\cite{wang22jacobi}), multiple frequency components will be scaled by the same $P_k(\lambda)$, thus affecting spectral filtering. Additionally, for each graph snapshot, linear \SpecTemGNNs cannot generate new frequency components if certain frequencies are missing from the original graph spectrum. For instance, in \shortautoref{fig: multidimensional and multivariate predictions}, a frequency component corresponding to $\lambda=1$ cannot be generated with a spectral filter. Although these two issues are challenging to address, they are rare in real-world attributed graphs \cite{wang22jacobi}.} \\

\noindent\textbf{Universal Temporal Spectral Filtering.} \revisiontwo{For a finite-length one-dimensional univariate time series, it can be modeled using a \textbf{\textit{frequency-domain model}} (FDM), consisting of sparse orthogonal space projectors and spectral filters. \shortautoref{fig: multidimensional and multivariate predictions} exemplifies modeling two time series with distinct TSFs. However, this assumes that the time series is well-represented in the transformed space. For instance, it may fail to fully capture non-stationary components or significant non-periodic trends that do not align with the chosen orthogonal projector, such as DFT. Further details are discussed in \shortautoref{subsec:design of spectral filters}.} \\

\noindent\textbf{Nonlinearity.} Nonlinear activation can be applied in both GSFs and TSFs. In the first case, we examine \revision{the role of element-wise nonlinearity over the spatial signal}, i.e., $\sigma(\mathbf{X}_t)$, enabling frequency components to be mixed w.r.t. the graph spectrum \cite{wang22jacobi}. In the second case, we investigate the role of nonlinearity over the temporal signal by studying its equivalent effect $\sigma'(\cdot)$, as $\sigma(\mathbf{X}_n) = \mathcal{F}^{-1}\big(\sigma'(\mathcal{F}(\mathbf{X}_n))\big)$. Here, we have $\sigma'(\Tilde{\mathbf{X}}_n) = \mathcal{F}\big(\sigma(\mathcal{F}^{-1}(\Tilde{\mathbf{X}}_n))\big)$, where different components in $\Tilde{\mathbf{X}}_n$ are first mixed (e.g., via \shortautoref{eq: inverse discrete fourier transform}) and then element-wise transformed by a nonlinear function $\sigma(\cdot)$ before being redistributed (e.g., via \shortautoref{eq: discrete fourier transform}). \revision{$\sigma'(\cdot)$ therefore functions as a column-wise nonlinear transformation over all frequency components. Consequently, a similar mixup exists, allowing different components to transform into each other in an orthogonal space, which may help mitigate issues such as missing frequency components from both spatial and temporal perspectives.} \\

\noindent\textbf{Connection to Dynamic Graph Isomorphism.} Analyzing the expressive power of GNNs is often laid on graph isomorphism. In this context, we first define the \textbf{\textit{temporal Weisfeiler-Lehman (WL) test}} and subsequently establish a connection to linear \SpecTemGNNs in \autoref{theorem: linear stgnn and wl-test}.

\begin{definition}[Temporal 1-WL test]\label{def: temporal 1-wl test} 
Temporal 1-WL test on discrete-time dynamic graphs $\mathcal{G}:=\{\mathcal{G}_t\}_{t=0}^{T-1}$ with a fixed node set $\mathbb{V}$ is defined below with iterative graph coloring procedures:

\vspace{1mm}
\quad\hangafter=1\hangindent=2.5em \tikzmarknode[mycircled,black]{a1}{1}\  \textit{\textbf{Initialization:}} All node colors are initialized using node features. In a snapshot at time $t$, we have $\forall v \in \mathbb{V}, c^{(0)}(v,t) = \mathbf{X}_{t}[v]$. In the absence of node features, all nodes get the same color.

\vspace{1mm}
\quad\hangafter=1\hangindent=2.5em \tikzmarknode[mycircled,black]{a1}{2}\  \textit{\textbf{Iteration:}} At step $l$, node colors are updated with an injective (hash) function: $\forall v \in \mathbb{V}, t \in [1,T), c^{(l+1)}(v,t) = \textsc{Hash}(c^{(l)}(v,t), c^{(l)}(v,t-1), \{\!\!\{ c^{(l)}(u,t) : e_{u,v,t} \in \mathbb{E}(\mathcal{G}_t) \}\!\!\})$. When $t=0$ or $T=1$, node colors are refined without $c^{(l)}(v,t-1)$ in the hash function.

\vspace{1mm}
\quad\hangafter=1\hangindent=2.5em \tikzmarknode[mycircled,black]{a1}{3}\  \textit{\textbf{Termination:}} The test is performed on two dynamic graphs in parallel, stopping when multisets of colors diverge at the end time, returning non-isomorphic. Otherwise, it is inconclusive.
\end{definition}

The temporal 1-WL test on DTDGs is an extension of the 1-WL test \cite{souza2022provably}.
Based on this, we demonstrate that the expressive power of linear \SpecTemGNNs is bounded by the temporal 1-WL test.

\begin{theorem}\label{theorem: linear stgnn and wl-test} 
For a linear \SpecTemGNN \revision{with valid temporal FDMs that can express arbitrary 1-dimensional univariate time series} and a $K$-degree polynomial basis in its GSFs, $\forall u,v \in \mathbb{V}, \mathbf{Z}_t[u] = \mathbf{Z}_t[v]$ if $\mathbf{C}^{(K+1)}(u,t) = \mathbf{C}^{(K+1)}(v,t)$. $\mathbf{Z}_t[i]$ and $\mathbf{C}^{(K)}(i,t)$ represent node $i$'s embedding at time $t$ in such a GNN and the $K$-step temporal 1-WL test, respectively.
\end{theorem}

\begin{figure}[t]
    \centering
    \includegraphics[width=0.99\linewidth]{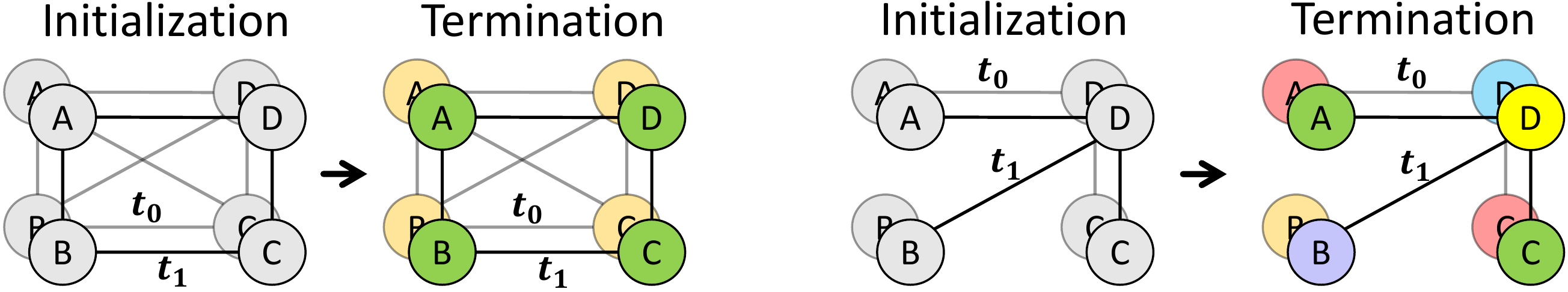}
    \caption{Two examples of temporal 1-WL test on non-attributive discrete-time dynamic graphs. The left test fails to distinguish non-isomorphic nodes at $t_1$, e.g., A and C, while the right example demonstrates a successful test.}
    \label{fig: temporal 1-wl examples}
\end{figure}

In other words, if a temporal 1-WL test cannot differentiate two nodes at a specific time, then a linear \SpecTemGNN will fail as well. However, this seems to contradict \autoref{theorem: linear stgnn}, where a linear \SpecTemGNN assigns any two nodes with different embeddings at a valid time step (under mild assumptions, regardless of whether they are isomorphic or not). For the temporal 1-WL test, it may not be able to differentiate some non-isomorphic temporal nodes and always assigns isomorphic nodes with the same representation/color. We provide examples in \shortautoref{fig: temporal 1-wl examples}. To resolve this discrepancy, we first prove that under mild conditions, i.e., a DTDG has no multiple eigenvalues and missing frequency components, the temporal 1-WL test can differentiate any non-isomorphic nodes at a valid time.

\begin{proposition}\label{prop: narrow temporal 1-wl down} 
If a discrete-time dynamic graph with a fixed graph topology at time $t$ has no repeated eigenvalues in its normalized graph Laplacian and has no missing frequency components in each snapshot, then the temporal 1-WL is able to differentiate all non-isomorphic nodes at time $t$.
\end{proposition}

We further prove that under the same assumptions, all pairs of nodes are non-isomorphic.

\begin{proposition}\label{prop: no graph automorphism} If a discrete-time dynamic graph with a fixed graph topology has no multiple eigenvalues in its normalized graph Laplacian and has no missing frequency components in each snapshot, then no automorphism exists.
\end{proposition}

Therefore, we close the gap and demonstrate that the expressive power of linear \SpecTemGNNs is theoretically bounded by the proposed temporal 1-WL test.

\subsection{Design of Spectral Filters}\label{subsec:design of spectral filters}
In this section, we outline a blueprint for designing powerful \SpecTemGNNs, with all proofs available in \textbf{\textit{Appendix A}}. We initially explore the optimal acquisition of spatial node embeddings within individual snapshots, focusing on the selection of the polynomial basis, $P_k(\cdot)$, for linear \SpecTemGNNs.

\begin{theorem}\label{theorem: design of spectral graph filters} For a linear \SpecTemGNN optimized with mean squared loss, any complete polynomial bases result in the same expressive power, but an orthonormal basis guarantees the maximum convergence rate if its weight function matches the graph signal density.
\end{theorem}

This theorem guides the design of GSFs for learning node embeddings in each snapshot w.r.t. the optimization of coefficients. Following this, we discuss the optimal modeling of temporal clues between snapshots in spectral domains. We begin by analyzing the function bases in space projectors.

\begin{lemma}\label{lemma: time series approximation} A time series with data points $x_j(t)$ can be expressed by $Q$ uncorrelated components $z_i(t)$ \revision{with an orthogonal (properly selected and possibly complex) projector~\cite{wallace1972empirical}}, i.e., $x_j(t) = \sum_{i=1}^{Q} e_{ij} z_i(t)$. The eigenvectors $e_i$ are orthogonal and determine the relationship between the data points $x_j(t)$.
\end{lemma}

\revision{By stabilizing the statistical properties of the input time series, DFT is robust and general enough to represent it in the spectral domain.} Additionally, operating on all spectral components is normally unnecessary~\cite{politransform,zhou22fedformer}. Consider a multivariate time series $\mathbf{X}_1(t), \cdots, \mathbf{X}_N(t)$, where each $T$-length univariate time series $\mathbf{X}_i(t)$ is transformed into a vector $\mathbf{a}_i = ({a}_{i,1}, \cdots, {a}_{i,T})^{\top} \in \mathbb{R}^{T \times 1}$ through a space projection. We form matrix $\mathbf{A} = (\mathbf{a}_1, \mathbf{a}_2, \cdots, \mathbf{a}_N)^{\top} \in \mathbb{R}^{N \times T}$ and apply a linear spectral filter as $\mathbf{A}\mathbf{W}$. Note that $\mathbf{A}$ does \textit{not} denote the adjacency matrix here. We then randomly select $S < T$ columns in $\mathbf{A}$ using the masking matrix $\mathbf{S} \in \{0,1\}^{S \times T}$, obtaining compact representation $\mathbf{A}' = \mathbf{A}\mathbf{S}^\top$ and linear spectral filtration as $\mathbf{A}'\mathbf{W}$. We demonstrate that, under mild conditions, $\mathbf{A}'\mathbf{W}$ preserves most information from $\mathbf{A}\mathbf{W}$. By projecting each column vector of $\mathbf{A}$ into the subspace spanned by column vectors in $\mathbf{A}'$, we obtain $P_{\mathbf{A}'}(\mathbf{A})=\mathbf{A}'(\mathbf{A}')^{\dagger}\mathbf{A}$. Let $\mathbf{A}_k$ represent $\mathbf{A}$'s approximation by its $k$ largest singular value decomposition. The lemma below shows $||\mathbf{A}\mathbf{W} - P_{\mathbf{A}'}(\mathbf{A})\mathbf{W}||_F$ is close to $||\mathbf{W}||_F||\mathbf{A} - \mathbf{A}_k||_F$ if the number of randomly sampled columns $S$ is on the order of $k^2$.

\begin{lemma}\label{lemma: linear filtering operator new} Suppose the projection of $\mathbf{A}$ by $\mathbf{A}'$ is $P_{\mathbf{A}'}(\mathbf{A})$, and the coherence measure of $\mathbf{A}$ is $\mu(\mathbf{A})=\Omega(k/N)$, then with a high probability, the error between $\mathbf{A}\mathbf{W}$ and $P_{\mathbf{A}'}(\mathbf{A})\mathbf{W}$ is bounded by $||\mathbf{A}\mathbf{W} - P_{\mathbf{A}'}(\mathbf{A})\mathbf{W}||_F \leq (1 + \epsilon)||\mathbf{W}||_F||\mathbf{A} - \mathbf{A}_k||_F$ if $S=O(k^2/\epsilon^2)$.
\end{lemma}

The lemmas above show that in most cases we can express a 1-dimensional time series with \textbf{(1)} orthogonal space projectors and \textbf{(2)} a reduced-order linear spectral filter. For practical application on $D$-dimensional multivariate time series data, we simply extend dimensions in $\mathbf{S}$ and $\mathbf{W}$.

\begin{theorem}\label{theorem: design of temporal frequency filters} Assuming accurate node embeddings in each snapshot, a linear \SpecTemGNN can, with high probability, produce expressive time series representations at valid times if its temporal FDMs consist of: \textbf{(1)} linear orthogonal space projectors; \textbf{(2)} individual reduced-order linear spectral filters.
\end{theorem}

\subsection{Connection to Related Work}\label{subsec: connection to related work}
Most of the deep time series models approximate expressive temporal filters with deep neural networks and learn important patterns directly in the time domain, e.g., TCN \cite{bai2018empirical}, where some complex properties (e.g., periodicity) may not be well modeled. Our work generalizes these methods in two ways: \textbf{(1)} when learning on univariate time series data, our temporal frequency-domain models guarantee that the most significant properties are well modeled with high probability (\autoref{lemma: time series approximation} and \autoref{lemma: linear filtering operator new}); \textbf{(2)} when learning on multivariate time series data, our framework models diverse inter-relations between time series (\shortautoref{fig: introduction signed graph}) and intra-relations within time series with theoretical evidence. 
Compared to \MPSTGNN, our framework has two major advantages: \textbf{(1)} it can model differently signed time series relations by learning a wide range of graph spectral filters (e.g., low-pass and high-pass), while MP-STGNNs only capture positive correlations between time series exhibiting strong similarities; \textbf{(2)} on this basis, it can express any multivariate time series under mild assumptions with provably expressive temporal frequency-domain models, while \MPSTGNN approximate effective temporal filtering in time domains with deep neural networks. Even compared to \SpaTemGNNs employing ChebyConv, our proposal generalizes them well: \textbf{(1)} we provide a blueprint for designing effective graph spectral filters and point out that using Chebyshev basis is sub-optimal; \textbf{(2)} instead of approximating expressive temporal models with deep neural networks, we detail how to simply construct them in frequency domains. Therefore, our results generalize most \SpaTemGNNs effectively. Although there are few studies on \SpecTemGNNs, such as StemGNN~\cite{cao2020spectral}, we are not only the first to define the general formulation and provide a theoretical framework to generalize this branch of methods but also free from the limitations of StemGNN mentioned above. Compared with spectral GNNs, such as BernNet~\cite{he2021bernnet} and JacobiConv~\cite{wang22jacobi}, our work extends graph convolution to model dynamic graphs comprising a sequence of regularly sampled graph snapshots.
\section{Methodology: Temporal Graph Gegenbauer Convolution}\label{sec:instant-tgc}
\begin{figure}[t]
    \centering
    \includegraphics[width=\linewidth]{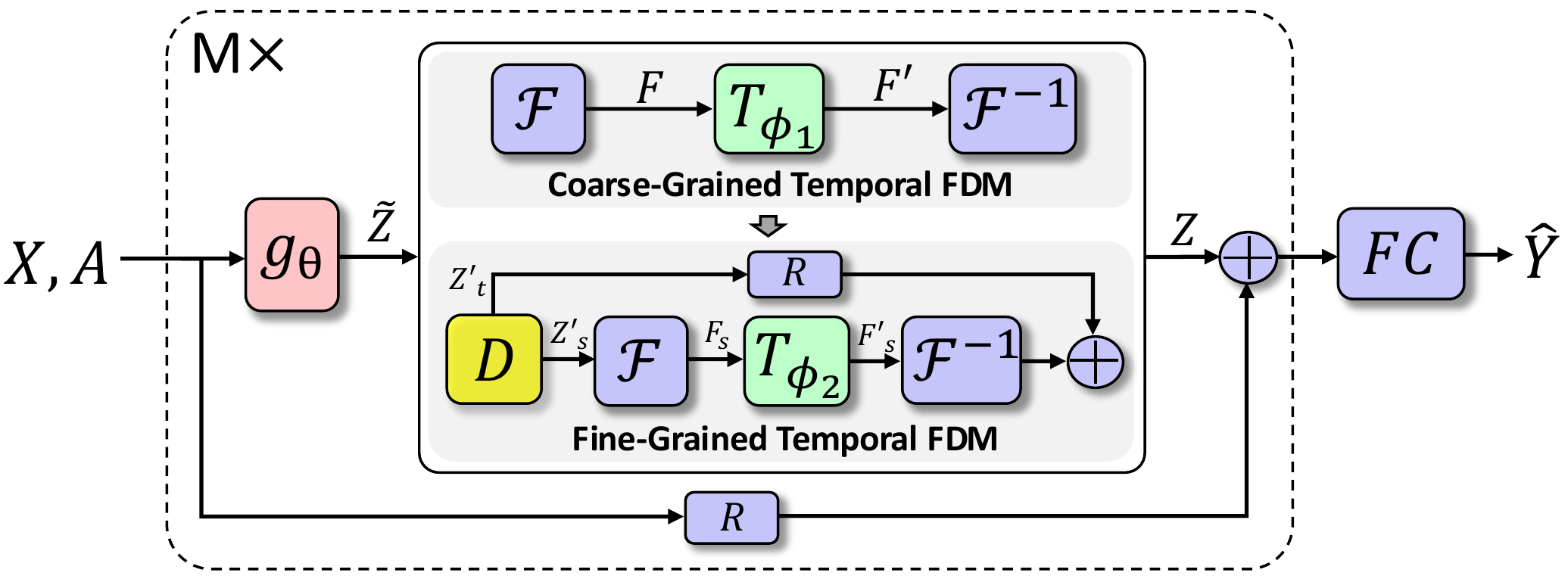}
    \caption{\revision{An illustration of Temporal Graph Gegenbauer Convolution (\TGC), where $g_\theta(\cdot)$, $\mathcal{T}_{\phi_1}(\cdot)$, and $\mathcal{T}_{\phi_2}(\cdot)$ are GSFs and two different TSFs. We use DFT and IDFT as $\mathcal{F}(\cdot)$ and $\mathcal{F}^{-1}(\cdot)$ in our implementation. See \shortautoref{fig: stgnn framework} for other notations.}}
    \label{fig:tgc_framework}
\end{figure}

In this section, we present a straightforward yet effective instantiation mainly based on the discussion in \shortautoref{subsec:design of spectral filters}. We first outline the basic formulation of the proposed \underline{\textbf{T}}emporal \underline{\textbf{G}}raph \underline{\textbf{G}}egenbauer \underline{\textbf{C}}onvolution (\TGC) and then connect it to other common practices. For the sake of clarity, we primarily present the canonical \TGC, which contains \textbf{\textit{only}} linear components in its building blocks and strictly inherits the basic formulation presented in \shortautoref{fig: stgnn framework}. Our primary aim here is not to achieve state-of-the-art performance, but rather to validate our theoretical insights through the examination of \TGC (and its advanced nonlienar variant, \TGCfull). \shortautoref{fig:tgc_framework} depicts the overall architecture of \TGC, which simply stacks $M$ building blocks as the pattern machine. Each block encompasses a graph convolution and two temporal frequency-domain models. \\

\noindent\textbf{Graph Gegenbauer Convolution.}
We implement $P_k(\cdot)$ in GSFs using the Gegenbauer basis due to its \textbf{(1)} generality and simplicity among orthogonal polynomials, \textbf{(2)} universality regarding its weight function, and \textbf{(3)} reduced model tuning expense. See the last paragraph in this section for a detailed justification. The Gegenbauer basis has the form $P_k^{\alpha}(x) = \frac{1}{k} [2x(k+ \alpha - 1)P_{k-1}^{\alpha}(x) - (k+ 2\alpha -2)P_{k-2}^{\alpha}(x)]$, with $P_0^{\alpha}(x) = 1$ and $P_1^{\alpha}(x) = 2 \alpha x$ for $k<2$. Specifically, $P_k^{\alpha}(x), k=0,1,...$ are orthogonal on the interval $[-1,1]$ w.r.t. the weight function $(1-x^2)^{\alpha-1/2}$. Based on this, we rewrite $P_k(\hat{\mathbf{L}})$ in \shortautoref{eq: general formula} as $P_k^{\alpha}(\mathbf{I} - \hat{\mathbf{L}}) = P_k^{\alpha}(\hat{\mathbf{A}})$, and the corresponding graph frequency-domain model (convolution) is defined as $\sum_{k=0}^{K} \theta_{k} P^{\alpha}_k(\hat{\mathbf{A}})\mathbf{X}$. \\

\noindent\textbf{Temporal Frequency-Domain Models.}
When designing temporal FDMs, linear orthogonal projections should be approximately sparse to support dimension reduction, e.g., DFT. For spectral filters, we randomly select $S$ frequency components before filtration. Specifically, for components $\mathbf{f} \in \mathbb{C}^{T}$ in $\mathbf{F} \in \mathbb{C}^{N \times T \times D}$ along $N$ and $D$ dimensions, we denote sampled components as $\mathbf{f}':= \mathbf{f}_{\mathbb{I}} \in \mathbb{C}^{S}$, where $\mathbb{I} = \{i_0, \cdots, i_{S-1}\}$ is a set of selection indices s.t. $\forall s \in \{0,\ldots,S-1\}$ and $i_{s-1} < i_{s}$. This is equivalent to $\mathbf{f}' = \mathbf{f}\Hat{\mathbf{S}}^\top$ with $\Hat{\mathbf{S}} \in \{0,1\}^{S \times T}$, where $\Hat{\mathbf{S}}_{i,s}=1$ if $i = i_s$. Thus, a standard reduced-order TSF is defined as $\mathbf{f}' = \mathbf{f}\Hat{\mathbf{S}}^\top\mathbf{W}$ with a trainable weight matrix $\mathbf{W} \in \mathbb{C}^{S \times S}$. \\

\begin{figure}[t]
    \centering
    \includegraphics[width=0.9\linewidth]{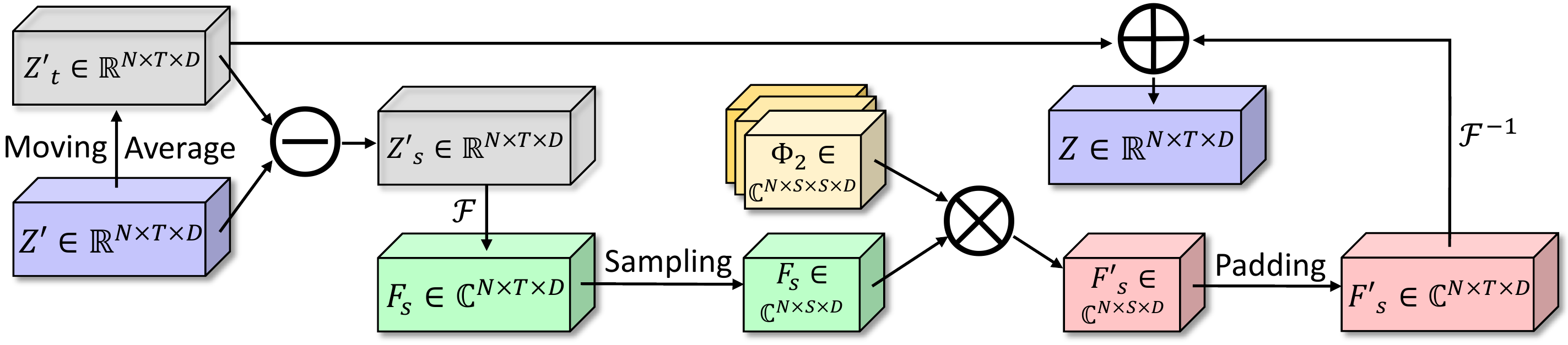}
    \caption{An illustration of tensor flow in fine-grained temporal frequency-domain models.}
    \label{fig:fdm_framework}
\end{figure}

\noindent\textbf{\TGC Building Block.}
Suppose $\Tilde{\mathbf{Z}} = \sum_{k=0}^{K} \theta_{k} P^{\alpha}_k(\hat{\mathbf{A}})\mathbf{X}$, we discuss two linear toy temporal FDMs. 
We first consider a basic coarse-grained filtering with the masking and weight matrices $\mathbf{S}_1$ and $\mathbf{\Phi}_1$:
\begin{equation}\label{eq:coarse-grained-temporal-frequency-filtering}
\begin{aligned}
\tikzmarknode[mycircled,black]{a1}{1}\ \mathbf{F} &= \mathcal{F}(\Tilde{\mathbf{Z}}),  &\hspace{0.3em}
\tikzmarknode[mycircled,black]{a1}{2}\ \mathbf{F}' &= \textsc{Pad}(\mathbf{F}\mathbf{S}_1^\top\mathbf{\Phi}_1), &\hspace{0.3em}
\tikzmarknode[mycircled,black]{a1}{3}\ \mathbf{Z}' &= \mathcal{F}^{-1}(\mathbf{F}').
\end{aligned}
\end{equation}
Next, we consider an \textbf{\textit{optional}} fine-grained filtering based on time series decomposition, where $\mathbf{S}_2$ and $\mathbf{\Phi}_2$ are optimized to further capture the information in detailed signals (e.g., seasonalities) while maintaining global time series profiles (i.e., trends). We formulize this process below and provide an illustration in \shortautoref{fig:fdm_framework}.
\begin{equation}\label{eq: fine-grained temporal frequency filtering}
\begin{aligned}
\tikzmarknode[mycircled,black]{a1}{1}\ \mathbf{Z}'_t, \mathbf{Z}'_s &= \textsc{Decomp}(\mathbf{Z}'), &\hspace{0.3em}
\tikzmarknode[mycircled,black]{a1}{2}\ \mathbf{F}_s &= \mathcal{F}(\mathbf{Z}'_s), &\hspace{0.3em} \\
\tikzmarknode[mycircled,black]{a1}{3}\ \mathbf{F}'_s &= \textsc{Pad}(\mathbf{F}_s \mathbf{S}_2^\top \mathbf{\Phi}_2), &\hspace{0.3em}
\tikzmarknode[mycircled,black]{a1}{4}\ \mathbf{Z} &= \mathbf{Z}'_t + \mathcal{F}^{-1}(\mathbf{F}'_s).
\end{aligned}
\end{equation}
After stacking multiple blocks, we forecast by transforming time series representations $\mathbf{Z}$.
For time series decomposition $\textsc{Decomp}(\cdot)$, we extract the trend information of a time series $\mathbf{x} \in \mathbb{R}^{T}$ by conducting moving average with a window size $w$ on the input signal, i.e., $ \Tilde{\mathbf{x}}(t) = \big(\mathbf{x}(t-w+1) + \cdots + \mathbf{x}(t)\big) / w$, where we simply pad the first $w-1$ data points with zeros in $\Tilde{\mathbf{x}}$. After this, we obtain detailed time series information (e.g., seasonality) by subtracting the trend from the input signal. For the component padding $\textsc{Pad}(\cdot)$, we pad the filtered signal tensors in both the coarse-grained and fine-grained temporal FDMs with $0+0j$, thereby reshaping them from $\mathbb{C}^{N \times S \times D}$ to $\mathbb{C}^{N \times T \times D}$. We illustrate this in \shortautoref{fig:fdm_framework}. \\

\begin{figure}[t]
    \centering
    \includegraphics[width=0.9\linewidth]{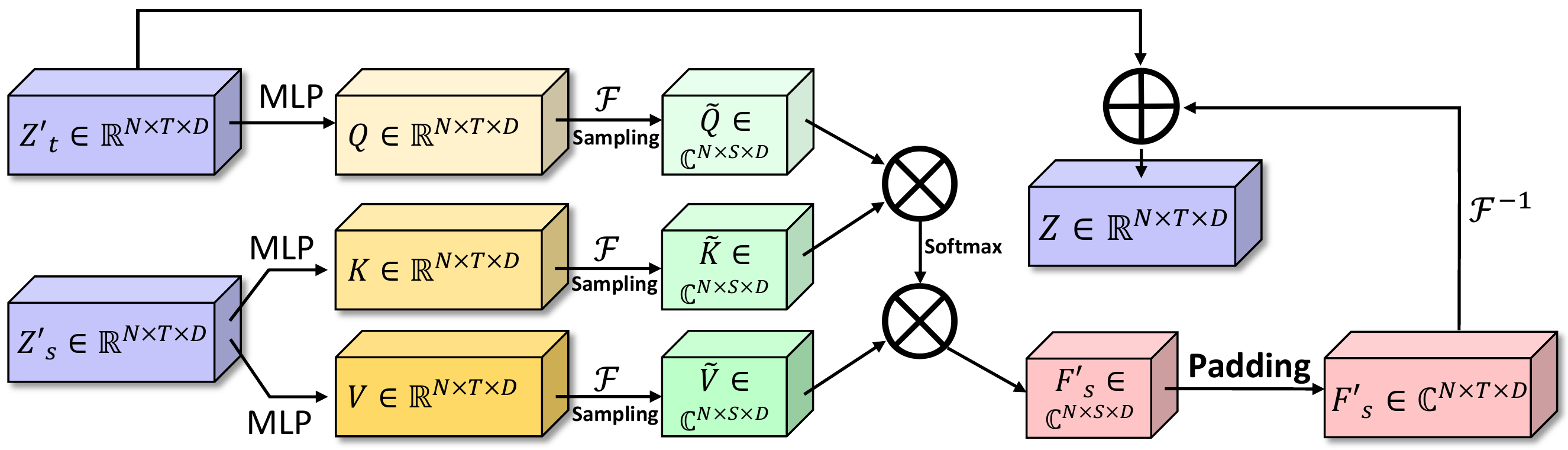}
    \caption{An illustration of tensor flow in fine-grained temporal frequency-domain models using spectral attention.}
    \label{fig:fdm_attention_framework}
\end{figure}

\noindent\textbf{Advanced Implementation.} Although the proposed \TGC model offers a straightforward and effective demonstration of the concepts, its performance can be enhanced by incorporating nonlinearities and other common modeling choices. \TGCfull is different from \TGC in two aspects: \textbf{(1)} we incorporate ReLU activation in both graph convolution and temporal FDMs, and \textbf{(2)} we implement the spectral filters in fine-grained temporal FDMs with the proposed spectral attention. 
\revision{An illustration is in \shortautoref{fig:fdm_attention_framework}.
We first decompose the input signal via $\mathbf{Z}'_t, \mathbf{Z}'_s = \textsc{Decomp}(\mathbf{Z}')$. After this, we generate the query matrix $\mathbf{Q} = \sigma(\mathbf{\Phi}_{2,1} \mathbf{Z}'_t) \in \mathbb{R}^{N \times T \times D}$, key matrix $\mathbf{K} = \sigma(\mathbf{\Phi}_{2,2} \mathbf{Z}'_s) \in \mathbb{R}^{N \times T \times D}$, and value matrix $\mathbf{V} = \sigma(\mathbf{\Phi}_{2,3} \mathbf{Z}'_s) \in \mathbb{R}^{N \times T \times D}$, where $\sigma(\cdot)$ denotes the ReLU activation, and $\mathbf{\Phi}_{2,i}$ are learnable weights. Then, we project all three matrices into the frequency domain with sparse components, i.e., $\Tilde{\mathbf{Q}} = \mathcal{F}(\mathbf{Q})\mathbf{S}^\top_{2,1} \in \mathbb{C}^{N \times S \times D}$, $\Tilde{\mathbf{K}} = \mathcal{F}(\mathbf{K})\mathbf{S}^\top_{2,2} \in \mathbb{C}^{N \times S \times D}$, and $\Tilde{\mathbf{V}} = \mathcal{F}(\mathbf{V})\mathbf{S}^\top_{2,3} \in \mathbb{C}^{N \times S \times D}$. Finally, we extract informative information with
\begin{equation}
    \mathbf{F}'_s = \textsc{Attention}(\Tilde{\mathbf{Q}}, \Tilde{\mathbf{K}}, \Tilde{\mathbf{V}}) = \textsc{Softmax}(\frac{\Tilde{\mathbf{Q}}\Tilde{\mathbf{K}}^\top}{\sqrt{n_q d_q}})\Tilde{\mathbf{V}},
    \label{eq: frequency attention}
\end{equation}
and we finally have $\mathbf{Z} = \mathbf{Z}'_t + \mathcal{F}^{-1}\big(\textsc{Pad}(\mathbf{F}'_s)\big)$.} \\

\begin{figure*}[t]
     \centering
     \begin{subfigure}[b]{0.22\linewidth}
         \centering
         \includegraphics[width=\linewidth]{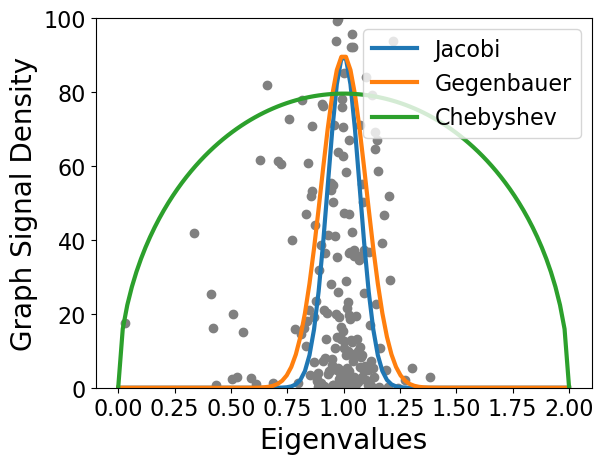}
         \caption{PeMS03 Dataset}
         \label{subfig: PeMS03 graph density}
     \end{subfigure}
     \hfill
     \begin{subfigure}[b]{0.22\linewidth}
         \centering
         \includegraphics[width=\linewidth]{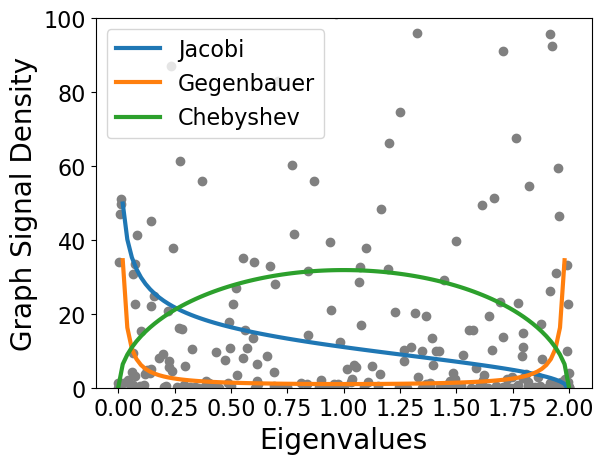}
         \caption{PeMS04 Dataset}
         \label{subfig: PeMS04 graph density}
     \end{subfigure}
     \hfill
     \begin{subfigure}[b]{0.22\linewidth}
         \centering
         \includegraphics[width=\linewidth]{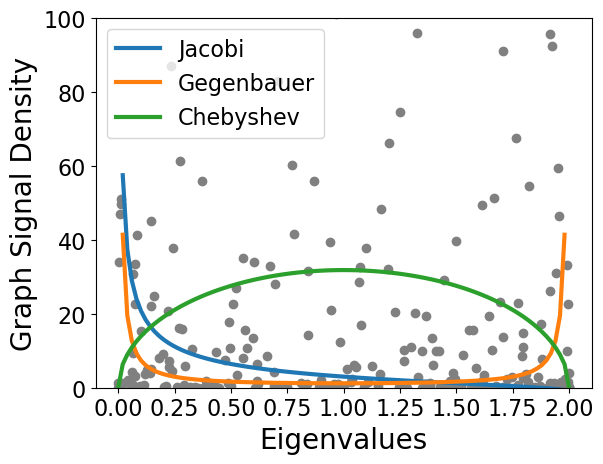}
         \caption{PeMS07 Dataset}
         \label{subfig: PeMS07 graph density}
     \end{subfigure}
     \hfill
     \begin{subfigure}[b]{0.22\linewidth}
         \centering
         \includegraphics[width=\linewidth]{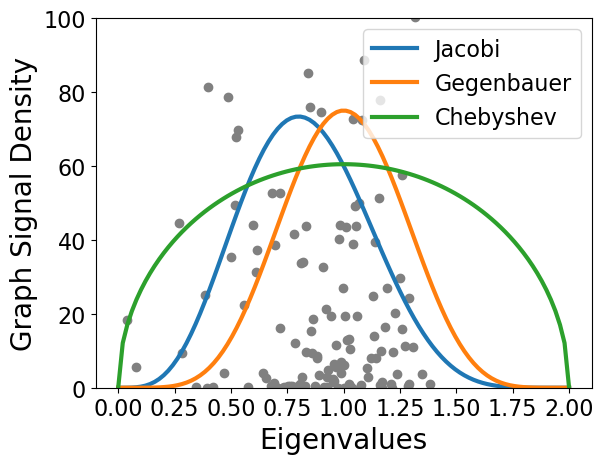}
         \caption{PeMS08 Dataset}
         \label{subfig: PeMS08 graph density}
     \end{subfigure}
        \caption{Signal density of predefined graphs on different datasets at a randomly selected time step versus the best-fitted weight functions of three different orthogonal polynomials.}
        \label{fig: signal density vs. weight function}
\end{figure*}

\noindent\textbf{Connection to Other Polynomial Bases.} 
We compare our design in \TGC with other common practices in approximating graph convolutions: Monomial, Chebyshev, Bernstein, and Jacobi bases. 
For non-orthogonal polynomials, our method with the Gegenbauer basis guarantees faster model convergences (\shortautoref{theorem: design of spectral graph filters}) and better empirical performances in most cases. Specifically, for the Monomial basis $(1-\lambda)^k$, it is non-orthogonal for arbitrary choices of weight functions \cite{wang22jacobi}. Although the Bernstein basis $\binom{K}{k}(1-\frac{\lambda}{2})^{K-k}(\frac{\lambda}{2})^{k}$ is also non-orthogonal, existing studies show that a small conditional number of the Hessian matrix $\kappa{(\mathbf{\mathbf{H}})}$ may still be achieved to enable fast convergence, where $\kappa{(\mathbf{\mathbf{H}})}$ can also be lower than using Monomial basis \cite{marco2010polynomial}. While the Bernstein basis is better than the Monomial basis in approximating graph convolutions, our implementation with the Gegenbauer basis guarantees the minimum $\kappa{(\mathbf{\mathbf{H}})}$ to be achieved in most cases; thus, it is more desired. We provide examples in \shortautoref{fig: signal density vs. weight function} showing that the weight functions of Gegenbauer polynomials fit graph signal densities well in most cases. We also confirm this in \shortautoref{tab: full ablation study}.
Compared with other orthogonal polynomials, we know that: \textbf{(1)} our basis is a generalization of the second-kind Chebyshev basis; \textbf{(2)} though our choice is a particular form of the Jacobi basis, the orthogonality of the Gegenbauer basis is well-posed in most real-world scenarios concerning its weight function. Particularly, the second-kind Chebyshev basis is a particular case of the Gegenbauer basis with $\alpha=1$ and only orthogonal w.r.t. a particular weight $\sqrt{1-\lambda^2}$. Though the Gegenbauer basis forms a particular case of the Jacobi basis with both of its parameters set to $\alpha-\frac{1}{2}$, we show that the orthogonality of the Gegenbauer basis is well-posed on common real-world graphs w.r.t. its weight function $(1-\lambda^2)^{\alpha-\frac{1}{2}}$ as shown in \shortautoref{fig: signal density vs. weight function}. Thus, we adopt the Gegenbauer basis as a simpler solution for our purpose with only minor performance degradation (\shortautoref{tab: full ablation study}).

\section{Evaluation}\label{sec:evaluation}
\begin{table}[h]
\centering
\caption{Statistics of eight different real-world time series datasets used in our work.}
\scalebox{0.65}{
\begin{tabular}{l|cccccccc}
\toprule
Statistic & PeMS03 & PeMS04 & PeMS07 & PeMS08 & Electricity & Solar  & Weather & ECG   \\ \hline
\# of time series & 358    & 307    & 228    & 170    & 321         & 137    & 21      & 140   \\
\# of data points & 26,209 & 16,992 & 12,672 & 17,856 & 26,304      & 52,560 & 52,696  & 5,000 \\
Sampling rate     & 5 min  & 5 min  & 5 min  & 5 min  & 1 hour      & 10 min & 10 min  & -     \\
Predefined graph  & Yes    & Yes    & Yes    & Yes    & No          & No     & No      & No  \\  \bottomrule
\end{tabular}}
\label{tab: dataset statistics}
\end{table}

In this section, we evaluate the effectiveness and efficiency of our proposal on 8 real-world datasets by comparing \TGC and \TGCfull with over 20 different baselines. To empirically validate our findings, we further perform extensive ablation studies and present a variety of visualizations using both synthetic and real-world examples.

\begin{table}[t]
\caption{Short-term forecasting results on four traffic benchmarks (\emph{Part I}). We use the \textcolor{red}{\textbf{bold}} and \textcolor{blue}{\underline{underline}} fonts to indicate the best and second-best results. We follow \cite{cao2020spectral} for the experimental setting when compared to \TGC.}
\label{tab:short-term-part1}
\scalebox{0.78}{
\begin{tabular}{p{1.8cm} *{8}{p{0.73cm}}}
\toprule
\multicolumn{1}{c}{\multirow{2}{*}{Method}} & MAE & RMSE         & MAE         & RMSE         & MAE         & RMSE         & MAE         & RMSE         \\ \cline{2-3} \cline{4-5} \cline{6-7} \cline{8-9}
\multicolumn{1}{c}{}                        & \multicolumn{2}{c}{\textit{PeMS03}} & \multicolumn{2}{c}{\textit{PeMS04}} & \multicolumn{2}{c}{\textit{PeMS07}} & \multicolumn{2}{c}{\textit{PeMS08}} \\ \hline
\textsc{LSTNet}~\cite{lai2018modeling}                                      &19.07             &29.67              &24.04             &37.38              &2.34             &4.26              &20.26             &31.69              \\
\textsc{DeepState}~\cite{rangapuram2018deep}                                   &15.59             &\textcolor{red}{\textbf{20.21}}              &26.50             &33.00              &3.95             &6.49              &19.34             &27.18              \\
\textsc{DeepGLO}~\cite{sen2019think}                                     &17.25             &23.25              &25.45             &35.90              &3.01             &5.25              &\textcolor{blue}{\underline{15.12}}             &25.22              \\
\textsc{DCRNN}~\cite{li2018diffusion}                                       &18.18             &30.31              &24.70             &38.12             &2.25             &4.04            &17.86             &27.83              \\
\textsc{STGCN}~\cite{yu2018spatio}                                      &17.49             &30.12              &22.70             &35.50              &2.25             &4.04              &18.02             &27.83              \\
\textsc{GWNet}~\cite{wu2019graph}                                      &19.85             &32.94              &26.85     &39.70              &-             &-              &19.13             &28.16              \\
\textsc{StemGNN}~\cite{cao2020spectral}                                     &\textcolor{blue}{\underline{14.32}}             &\textcolor{blue}{\underline{21.64}}              &\textcolor{blue}{\underline{20.24}}             &\textcolor{blue}{\underline{32.15}}              &\textcolor{blue}{\underline{2.14}}             &\textcolor{blue}{\underline{4.01}}
&15.83             &\textcolor{blue}{\underline{24.93}}              \\           
\rowcolor{tabhighlight} \textsc{\TGC} \textbf{(Ours)} & \textcolor{red}{\textbf{13.52}}             &21.74              &\textcolor{red}{\textbf{18.77}}          &\textcolor{red}{\textbf{29.92}}              &\textcolor{red}{\textbf{1.92}}             &\textcolor{red}{\textbf{3.35}}              &\textcolor{red}{\textbf{14.55}}             &\textcolor{red}{\textbf{22.73}}\\       
\bottomrule
\end{tabular}
}
\end{table}
\vspace{-2mm}
\revisiontwo{\subsection{Experiment Details}}
\noindent\textbf{Datasets.} For short-term forecasting, we evaluate on 3 time series and 4 traffic datasets commonly used as benchmarks for forecasting models. The PeMS03, 04, 07, and 08 datasets are derived from traffic sensors in highway systems throughout California's metropolitan areas~\footnote{\url{https://pems.dot.ca.gov/}}. They are sampled every 5 minutes. 
The Electricity dataset~\footnote{\url{https://github.com/laiguokun/multivariate-time-series-data}\label{fn: electricity dataset}} contains electricity consumption data for 321 customers, sampled hourly. The Weather dataset~\footnote{\url{https://github.com/zhouhaoyi/Informer2020}} features the one-year records of 21 meteorological stations installed in Germany, sampled every 10 minutes. The ECG dataset includes 5,000 records from 140 electrocardiograms in the URC time series archive~\footnote{\url{https://www.cs.ucr.edu/~eamonn/time_series_data/}}. For long-term forecasting, we utilize the Electricity and Weather datasets, alongside an additional Solar-Energy dataset\textsuperscript{\ref{fn: electricity dataset}}. This dataset consists of photovoltaic production from 137 sites in Alabama in 2006, with a 10-minute sampling rate. \\

\noindent\textbf{Baselines.} We compare the proposed methods with a list of representative and competitive baselines across different datasets and tasks. In short-term forecasting, our baselines include several deep time series methods: FC-LSTM~\cite{sutskever2014sequence}, TCN~\cite{bai2018empirical}, LSTNet~\cite{lai2018modeling}, DeepState~\cite{rangapuram2018deep}, DeepGLO~\cite{sen2019think}, and SFM~\cite{zhang2017stock}. We also compare with a series of spatio-temporal graph neural networks: DCRNN~\cite{li2018diffusion}, STGCN~\cite{yu2018spatio}, Graph WaveNet~\cite{wu2019graph}, ASTGCN~\cite{guo2019attention}, STG2Seq~\cite{bai2019stg2seq}, LSGCN~\cite{huang2020lsgcn}, STSGCN~\cite{sofianos2021space}, STFGNN~\cite{li2021spatial}, STGODE~\cite{fang2021spatial}, and StemGNN~\cite{cao2020spectral}. In long-term forecasting, we further compare with several state-of-the-art Transformer-based models: FiLM~\cite{zhou2022film}, FEDformer~\cite{zhou22fedformer}, Autoformer~\cite{wu2021autoformer}, Informer~\cite{zhou2021informer}, LogTrans~\cite{li2019enhancing}, and Reformer~\cite{kitaev2019reformer}. \\

\noindent\textbf{Implementation Details.} All experiments are conducted on 4 $\times$ Nvidia RTX 2080 Ti 11GB GPUs. We utilized the Mean Absolute Forecasting Errors (MAE) and Root Mean Squared Forecasting Errors (RMSE) as our primary metrics. Our results are averaged over 5 runs.

\begin{table}[t]
\caption{Short-term forecasting results on four traffic benchmarks (\emph{Part II}). We use the \textcolor{red}{\textbf{bold}} and \textcolor{blue}{\underline{underline}} fonts to indicate the best and second-best results. We follow \cite{choi2022graph} for the experimental setting when compared to \TGCfull.}
\label{tab:short-term-part2}
\scalebox{0.78}{
\begin{tabular}{p{1.9cm} *{8}{p{0.73cm}}}
\toprule
\multicolumn{1}{c}{\multirow{2}{*}{Method}} & MAE& RMSE         & MAE         & RMSE         & MAE         & RMSE         & MAE         & RMSE         \\ \cline{2-3} \cline{4-5} \cline{6-7} \cline{8-9}
\multicolumn{1}{c}{}                        & \multicolumn{2}{c}{\textit{PeMS03}} & \multicolumn{2}{c}{\textit{PeMS04}} & \multicolumn{2}{c}{\textit{PeMS07}} & \multicolumn{2}{c}{\textit{PeMS08}} \\ \hline
\textsc{ASTGCN}~\cite{guo2019attention}                                      &17.34             &29.56              &22.93             &35.22              &3.14             &6.18              &18.25             &28.06              \\
\textsc{MSTGCN}~\cite{guo2019attention}                                   &19.54             &31.93              &23.96             &37.21              &3.54             &6.14              &19.00             &29.15              \\
\textsc{STG2Seq}~\cite{bai2019stg2seq}                                     &19.03             &29.83              &25.20             &38.48              &3.48             &6.51              &20.17             &30.71              \\
\textsc{LSGCN}~\cite{huang2020lsgcn}                                       &17.94             &29.85              &21.53             &33.86             &3.05             &5.98           &17.73             &26.76              \\
\textsc{STSGCN}~\cite{sofianos2021space}                                      &17.48             &29.21              &21.19             &33.65              &3.01             &5.93              &17.13            &26.80              \\
\textsc{STFGNN}~\cite{li2021spatial}                                      &16.77             &28.34              &\textcolor{blue}{\underline{20.48}}     &\textcolor{blue}{\underline{32.51}}              &\textcolor{blue}{\underline{2.90}}             &5.79              &16.94             &26.25              \\
\textsc{STGODE}~\cite{fang2021spatial}                                     &\textcolor{blue}{\underline{16.50}}             &\textcolor{blue}{\underline{27.84}}              &20.84             &32.82              &2.97             &\textcolor{blue}{\underline{5.66}}        &\textcolor{blue}{\underline{16.81}}             &\textcolor{red}{\textbf{25.97}}              \\           
\rowcolor{tabhighlight} \textsc{\TGCfull} \textbf{(Ours)} & \textcolor{red}{\textbf{16.22}}             &\textcolor{red}{\textbf{27.07}}              &\textcolor{red}{\textbf{20.00}}          &\textcolor{red}{\textbf{32.10}}              &\textcolor{red}{\textbf{2.81}}             &\textcolor{red}{\textbf{5.58}}              &\textcolor{red}{\textbf{16.54}}             &\textcolor{blue}{\underline{26.10}} \\       
\bottomrule
\end{tabular}
}
\end{table}
\begin{table}[t]
\centering
\caption{Efficiency comparison of representative models: Trainable parameters (M), time-per-epoch (s), and total training time (min); $\diamondsuit$ indicates significantly larger values compared to other methods.}
\resizebox{1.0\linewidth}{!}{
\begin{tabular}{p{43 pt}p{34 pt}<{\centering}p{34 pt}<{\centering}p{34 pt}<{\centering}p{34 pt}<{\centering}}
\toprule
Method    & \textit{PeMS03} & \textit{PeMS04} & \textit{PeMS07} & \textit{PeMS08} \\ \hline
\textsc{LSTNet}    &  0.4/2.2/3.8    &  0.3/1.3/2.3    & 0.2/0.9/1.5     & 
0.2/1.0/1.7     \\
\textsc{DeepGLO} & 0.6/14/8.3                & 0.6/8.5/6.0                &0.3/6.2/6.6                 &0.3/5.9/4.2                 \\
\textsc{DCRNN}     &{OOM}                 &0.4/$\diamondsuit$/$\diamondsuit$                 &0.4/$\diamondsuit$/$\diamondsuit$                &0.4/$\diamondsuit$/$\diamondsuit$                 \\
\textsc{STGCN}    &0.3/25/13                 &0.3/14/6.9                 &0.2/8.6/4.1                 &0.2/8.0/5.0                 \\
\textsc{StemGNN}   &1.4/17/24                 &1.3/9.0/13                 &1.2/6.0/8.0                 &1.1/6.0/9.0                \\
\rowcolor{tabhighlight} \TGC \tiny{\textbf{(Ours)}}      &0.4/12/21                &0.3/6.2/11                 &0.2/3.9/7.2                 &0.1/4.2/8.4                \\ 
\bottomrule
\end{tabular}}
\label{tab: model efficiency}
\end{table}

\begin{table*}[t]
\small
\centering
\caption{Long-term forecasting results on three time series benchmarks. We follow \cite{zhou2022film} for the experimental setting when compared to \TGCfull.}
\setlength\extrarowheight{3.5pt}
\scalebox{0.8}{
\begin{tabular}{p{10 pt} p{5 pt} | >{\columncolor{tabhighlight}}p{30 pt}>{\columncolor{tabhighlight}} p{30 pt} p{30 pt} p{30 pt} p{30 pt} p{30 pt} p{30 pt} p{30 pt} p{30 pt} p{30 pt} p{30 pt} p{30 pt} p{30 pt} p{30 pt}}
\toprule
\multicolumn{2}{c|}{Method}        & \multicolumn{2}{c|}{\textsc{\TGCfull} \small{\textbf{(Ours)}}} & \multicolumn{2}{c|}{\textsc{FiLM}~\cite{zhou2022film}} & \multicolumn{2}{c|}{\textsc{FEDformer}~\cite{zhou22fedformer}} & \multicolumn{2}{c|}{\textsc{Autoformer}~\cite{wu2021autoformer}} & \multicolumn{2}{c|}{\textsc{Informer}~\cite{zhou2021informer}} & \multicolumn{2}{c|}{\textsc{LogTrans}~\cite{li2019enhancing}} & \multicolumn{2}{c}{\textsc{Reformer}~\cite{kitaev2019reformer}} \\ \hline
\multicolumn{2}{c|}{Metric}        & MAE         & RMSE        & MAE           & RMSE           & MAE            & RMSE           & MAE           & RMSE          & MAE           & RMSE          & MAE          & RMSE
& MAE            & RMSE
\\ \hline
\multicolumn{1}{c|}{\multirow{3}{*}{\rotatebox{90}{\small{\textit{Electricity}}}}} & \multicolumn{1}{c|}{96}  &\textcolor{blue}{\underline{0.293}}            &\textcolor{blue}{\underline{0.425}}            &\textcolor{red}{\textbf{0.267}}               &\textcolor{red}{\textbf{0.392}}                &0.297                &0.427                &0.317               &0.448               &0.368               &0.523               &0.357              &0.507               &0.402    &0.558   \\
\multicolumn{1}{c|}{}                             & \multicolumn{1}{c|}{192} &\textcolor{blue}{\underline{0.303}}            &\textcolor{blue}{\underline{
0.440}}          &\textcolor{red}{\textbf{0.258}}               &\textcolor{red}{\textbf{0.404}}                &0.308                &0.442                &0.334               &0.471               &0.386               &0.544               &0.368              &0.515               &0.433                   &0.590 \\
\multicolumn{1}{c|}{}                             & \multicolumn{1}{c|}{336} &\textcolor{blue}{\underline{0.313}}             &0.470             & \textcolor{red}{\textbf{0.283}}               &\textcolor{red}{\textbf{0.433}}                &0.313                &\textcolor{blue}{\underline{0.460}}                &0.338              &0.480               &0.394               &0.548               &0.380              &0.529               &0.433                   &0.591\\ \hline
\multicolumn{1}{c|}{\multirow{3}{*}{\rotatebox{90}{\small{\textit{Weather}}}}}     & \multicolumn{1}{c|}{96}  &\textcolor{red}{\textbf{0.235}}             &\textcolor{red}{\textbf{0.408}}             &\textcolor{blue}{\underline{0.262}}               &\textcolor{blue}{\underline{0.446}}                &0.296                &0.465                &0.336             &0.515               &0.384               &0.547               &0.490              &0.677               &0.596                 &0.830 \\
\multicolumn{1}{c|}{}                             & \multicolumn{1}{c|}{192} &\textcolor{red}{\textbf{0.286}}             &\textcolor{red}{\textbf{0.468}}              &\textcolor{blue}{\underline{0.288}}               &\textcolor{blue}{\underline{0.478}}                & 0.336                &0.525                &0.367               &0.554               &0.544               &0.773               &0.589              &0.811               &0.638                   &0.867 \\
\multicolumn{1}{c|}{}                             & \multicolumn{1}{c|}{336} &\textcolor{red}{\textbf{0.317}}             &\textcolor{red}{\textbf{0.515}}             &\textcolor{blue}{\underline{0.323}}               &\textcolor{blue}{\underline{0.516}}              &0.380                &0.582                &0.395               &0.599               &0.523               &0.760               &0.652              &0.892               &0.596           &0.799   \\ \hline
\multicolumn{1}{c|}{\multirow{3}{*}{\rotatebox{90}{\small{\textit{Solar}}}}}    & \multicolumn{1}{c|}{96}  & \textcolor{red}{\textbf{0.242}}            & \textcolor{red}{\textbf{0.443}}            &0.311               &0.557               &0.363            &0.448                &0.552               &0.787               &0.264               &0.469               &0.262              &0.467               &\textcolor{blue}{\underline{0.255}}         &\textcolor{blue}{\underline{0.451}}       \\
\multicolumn{1}{c|}{}                             & \multicolumn{1}{c|}{192} &\textcolor{red}{\textbf{0.263}}            &\textcolor{red}{\textbf{0.470}}            &0.356               &0.595                &0.354                 &0.483               &0.674               &0.856               &0.280               &0.487               &0.284              &0.489               &\textcolor{blue}{\underline{0.274}}               &\textcolor{blue}{\underline{0.475}}   \\
\multicolumn{1}{c|}{}                             & \multicolumn{1}{c|}{336} &\textcolor{red}{\textbf{0.271}}            &\textcolor{red}{\textbf{0.478}}            &0.370          &0.628                &0.372               & 0.518                &0.937               &1.131               &0.285               &0.496               &0.295              &0.512               &\textcolor{blue}{\underline{0.278}}               &\textcolor{blue}{\underline{0.491}}   \\ 
\bottomrule
\end{tabular}}
\label{tab: long-term forecasting}
\end{table*}
\begin{figure*}[t]
     \centering
     \begin{subfigure}[b]{0.49\linewidth}
         \centering
         \includegraphics[width=\linewidth]{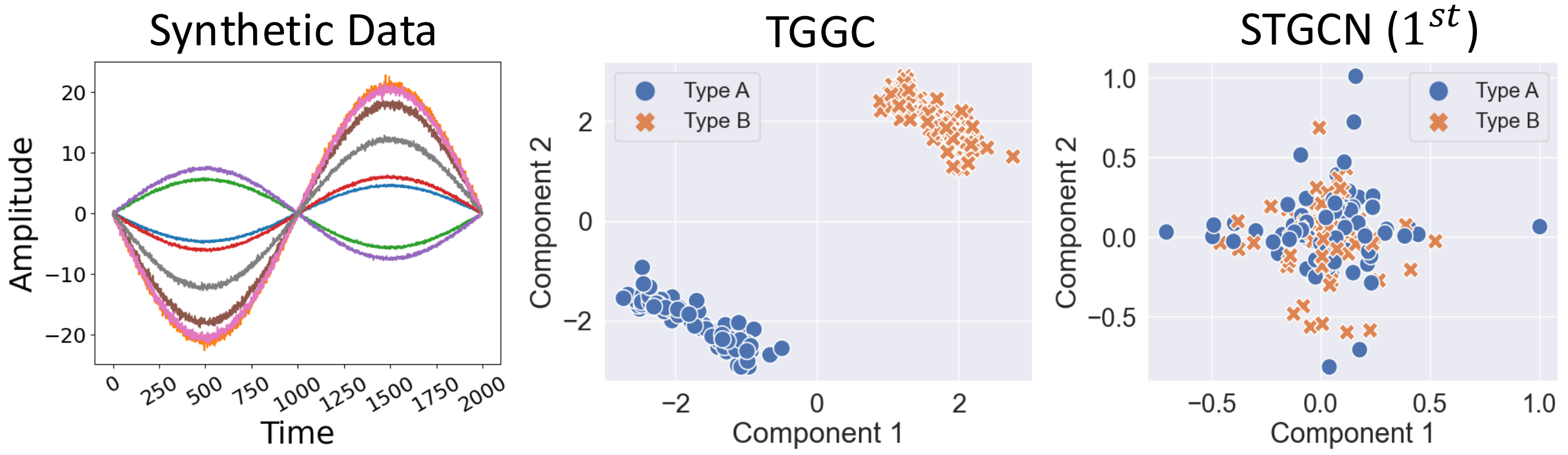}
         \caption{Visualization of learned embeddings w.r.t. different time series correlations on a synthetic dataset. Types A and B represent series groups with opposing trends (e.g., pink vs. green).}
         \label{subfig: signed relation learning experiments on synthetic data}
     \end{subfigure}
     \hfill
     \begin{subfigure}[b]{0.49\linewidth}
         \centering
         \includegraphics[width=\linewidth]{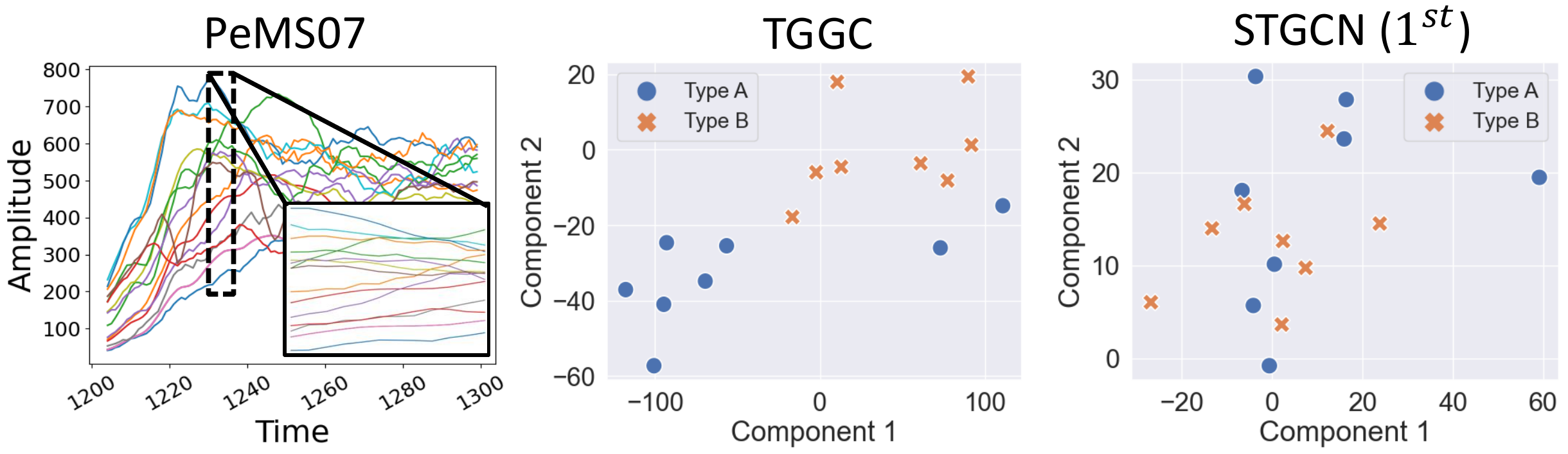}
         \caption{Visualization of learned embeddings w.r.t. different time series correlations on PeMS07 dataset. Types A and B represent series groups with opposing trends.}
         \label{subfig: signed relation learning experiments on pems07 data}
     \end{subfigure}
        \caption{Evaluation of learning differently signed time series relations.}
        \label{fig: signed relation learning experiments}
\end{figure*}

\begin{figure}[t]
    \centering
    \includegraphics[width=0.99\linewidth]{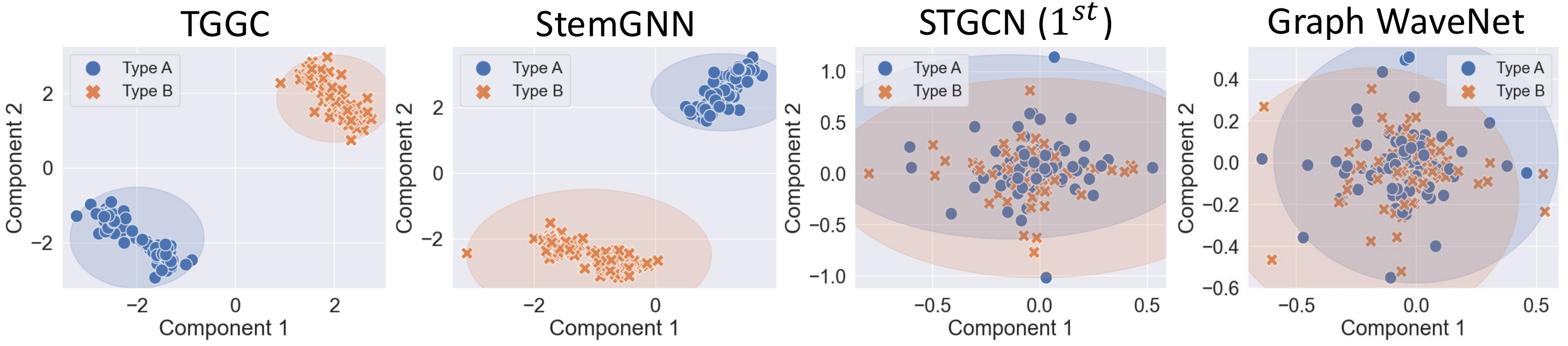}
    \vfill
    \includegraphics[width=0.99\linewidth]{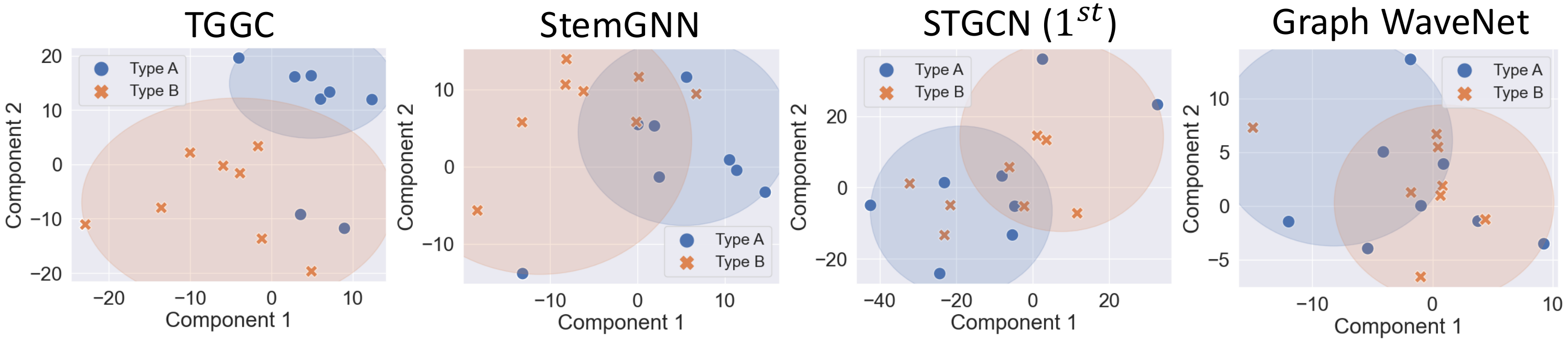}
    \caption{Additional comparison between learned embeddings w.r.t. different time series correlations on a synthetic dataset presented in \shortautoref{subfig: signed relation learning experiments on synthetic data} \textbf{(upper)} and a subset of traffic recordings in PeMS07 dataset as in \shortautoref{subfig: signed relation learning experiments on pems07 data} \textbf{(lower)}. Types A and B denote series groups exhibiting opposing trends. Differently colored shading indicates distinct clusters.}
    \label{fig: additional embedding visualization}
\end{figure}
For short-term forecasting, our default data splits are 60\%-20\%-20\% for the PeMS03, 04, and 08 datasets, and 70\%-20\%-10\% for the others as per \cite{cao2020spectral}. Min-max normalization is applied to the ECG dataset, while Z-Score normalization is employed for the rest. Using the past 1-hour observations from four traffic datasets, we predict the next 15-minute traffic volumes or speeds, a setting consistently adopted in ablation studies and model efficiency comparisons. For the Electricity dataset, the prior 24-hour readings forecast the subsequent 1-hour consumption. The Solar-Energy dataset employs the previous 4 hours of data to predict the next half-hour production. The ECG dataset has a window size of 12 and a forecasting horizon of 3. All baseline configurations are the same as in \cite{cao2020spectral}. In \shortautoref{tab:short-term-part2}, we compare to a series of competitive STGNNs and follow the setup in \cite{choi2022graph} with minor differences: we use the 60\%-20\%-20\% data split across all four datasets, and leverage the past 1-hour observations to predict the next 1-hour traffic volume or speed.

For long-term forecasting, we adhere to \cite{zhou2022film}, using a 70\%-10\%-20\% split for all datasets. The Electricity dataset uses 4 days of past observations to predict the next 4, 8, and 14 days. Both the Solar-Energy and Weather datasets employ 16-hour past data to predict the subsequent 16, 32, and 56 hours. Hyperparameter settings are detailed in \textbf{\textit{Appendix B}}.

\vspace{2mm}
\subsection{Main Results.}
We evaluate against related work in terms of model effectiveness (\shortautoref{tab:short-term-part1}, \shortautoref{tab:short-term-part2}, and \shortautoref{tab: long-term forecasting}) and efficiency (\shortautoref{tab: model efficiency}), showcasing the potential of \SpecTemGNNs for time series forecasting. 
We compare our vanilla instantiation (\TGC) with the most pertinent and representative works in \shortautoref{tab:short-term-part1} on four traffic benchmarks. Next, we assess our method (\TGCfull, the nonlinear version of \TGC) against a series of competitive STGNNs in \shortautoref{tab:short-term-part2}. 
The main differences here are the data splits and forecasting horizons we followed in \cite{cao2020spectral} and \cite{choi2022graph} as mentioned above.
Additionally, we report long-term forecasting results in \shortautoref{tab: long-term forecasting}, comparing our approach with state-of-the-art Transformer models.

Our method consistently outperforms most baselines by significant margins. In short-term forecasting, \TGC and \TGCfull achieve average performance gains of \textbf{7.3\%} and \textbf{1.7\%} w.r.t. the second-best results. Notably, we observe a significant improvement ($\sim$\textbf{8\%}) over StemGNN~\cite{cao2020spectral}, a special case of our method with nonlinearities. Considerable enhancements are also evident when compared to ASTGCN~\cite{guo2019attention} ($\sim$\textbf{9\%}) and LSGCN~\cite{huang2020lsgcn} ($\sim$\textbf{7\%}), which primarily differ from StemGNN in temporal dependency modeling and design nuances. In long-term forecasting, our method further exhibits impressive performance, outperforming the second-best results by about \textbf{3.3\%}. These results indicate that even simple, yet appropriately configured \SpecTemGNNs (discussed in \shortautoref{subsec:design of spectral filters}) are potent time series predictors. In \shortautoref{tab: model efficiency}, we examine our method's efficiency by comparing \TGC to representative baselines. We find that \TGC forms the simplest and most efficient \SpecTemGNN to date compared with StemGNN.  In comparison to other \SpaTemGNNs, such as STGCN~\cite{yu2018spatio} and DCRNN~\cite{li2018diffusion}, our method also exhibits superior model efficiency across various aspects. Though time series models like LSTNet~\cite{lai2018modeling} are faster in training, they do not have spatial modules and thus less effective.

\begin{table*}[htbp]
\centering
\setlength\extrarowheight{2pt}
\caption{Ablation study results. We use the \textcolor{red}{\textbf{bold}} and \textcolor{blue}{\underline{underline}} fonts to denote the best and second-best results in each ablation block, respectively.}
\scalebox{0.99}{
\begin{tabular}{p{3.1cm} *{8}{p{1.3cm}}}
\toprule
\multicolumn{1}{c}{\multirow{2}{*}{Variant}} & MAE         & RMSE         & MAE         & RMSE         & MAE         & RMSE         & MAE         & RMSE         \\ \cline{2-3} \cline{4-5} \cline{6-7} \cline{8-9}
\multicolumn{1}{c}{}                        & \multicolumn{2}{c}{\textit{PeMS03}} & \multicolumn{2}{c}{\textit{PeMS04}} & \multicolumn{2}{c}{\textit{PeMS07}} & \multicolumn{2}{c}{\textit{PeMS08}} \\ \hline
\textbf{A.1} Monomial                                    &27.64             &43.52              &59.41            &120.18              &5.68             &8.73              &29.36             &43.37              \\
\textbf{A.2} Bernstein                                   &27.38             &43.17              &55.17             &105.13              &5.57             &8.64              &27.57             &40.28              \\
\textbf{A.3} Chebyshev                                   &13.56             &21.84              &18.78             &\textcolor{blue}{\underline{29.89}}             &1.94             &3.37              &14.36             &22.93              \\
\textbf{A.4} Gegenbauer                                  &\textcolor{blue}{\underline{13.52}}            & \textcolor{blue}{\underline{21.74}}            &\textcolor{blue}{\underline{18.77}}             &29.92              &\textcolor{blue}{\underline{1.92}}            &\textcolor{blue}{\underline{3.35}}              &\textcolor{blue}{\underline{14.35}}             &\textcolor{blue}{\underline{22.73}}             \\
\textbf{A.5} Jacobi                                      &\textcolor{red}{\textbf{13.14}}             &\textcolor{red}{\textbf{21.51}}              &\textcolor{red}{\textbf{18.64}}            &\textcolor{red}{\textbf{29.44}}          &\textcolor{red}{\textbf{1.91}}            &\textcolor{red}{\textbf{3.34}}              &\textcolor{red}{\textbf{14.29}}             &\textcolor{red}{\textbf{22.15}}              \\ \hline\hline
\rowcolor{tabhighlight}\TGC \textbf{(Ours)}                                      &\textcolor{red}{\textbf{13.52}}             &21.74             &\textcolor{red}{\textbf{18.77}}             &\textcolor{red}{\textbf{29.92}}              &\textcolor{red}{\textbf{1.92}}             &\textcolor{red}{\textbf{3.35}}              &\textcolor{red}{\textbf{14.55}}             &\textcolor{red}{\textbf{22.73}}              \\
\textbf{B.1} w/o MD-F                                  &14.07             &21.82              &19.07             &30.34              &2.05             &3.45              &15.14            & 23.49              \\
\textbf{B.2} w/o MD-F$^\dagger$                          &\textcolor{blue}{\underline{13.62}}             &21.73              &18.92             &30.11              &1.96             &3.40              &15.06             & 23.25             \\
\textbf{B.3} w/o MV-F                         &13.80             &21.77              &19.12             &30.50              &1.97             &3.44              &15.12             &23.58              \\
\textbf{B.4} w/o O-SP                         &13.72             &21.75              &19.10             &30.42              &2.03             &3.53              &15.38            &28.84              \\
\textbf{B.5} w/o C-FDM                                  &13.83             &\textcolor{red}{\textbf{21.47}}              &19.15             &30.52              &2.01             &3.47              &15.18             &23.71              \\
\textbf{B.6} w/o F-FDM                                  &13.71             &\textcolor{blue}{\underline{21.59}}              &\textcolor{blue}{\underline{18.81}}             &\textcolor{blue}{\underline{29.96}}              &\textcolor{blue}{\underline{1.93}}             &\textcolor{blue}{\underline{3.39}}              &\textcolor{blue}{\underline{14.88}}             &\textcolor{blue}{\underline{23.10}}              \\

\hline
\rowcolor{tabhighlight}\TGCfull\textbf{(Ours)} &\textcolor{red}{\textbf{13.39}}            &\textcolor{red}{\textbf{21.34}}              &\textcolor{red}{\textbf{18.41}}          &\textcolor{red}{\textbf{29.39}}              &\textcolor{red}{\textbf{1.84}}             &\textcolor{red}{\textbf{3.28}}              &\textcolor{red}{\textbf{14.38}}             &\textcolor{red}{\textbf{22.43}} \\ 
\textbf{C.1} w/o NL                                        &13.75             &21.43              &18.63             &29.70              &1.92             &3.36              &14.94             &24.93              \\
\textbf{C.2} w/o S-Attn                                        &\textcolor{blue}{\underline{13.42}}             &\textcolor{blue}{\underline{21.38}}              &\textcolor{blue}{\underline{18.50}}             &\textcolor{blue}{\underline{29.53}}              &\textcolor{blue}{\underline{1.86}}             &\textcolor{blue}{\underline{3.30}}              &\textcolor{blue}{\underline{14.43}}             &\textcolor{blue}{\underline{22.54}}            \\ 
\bottomrule
\end{tabular}}
\label{tab: full ablation study}
\end{table*}

\subsection{Evaluation of Modeling Time Series Dependencies.} 
Our method, in contrast to most GNN-based approaches, excels at learning different signed spatial relations between time series (\shortautoref{fig: introduction signed graph}). Predetermined or learned graph topologies typically reflect the strength of underlying connectivity, yet strongly correlated time series might exhibit distinct properties (e.g., trends), which \MPSTGNN often struggles to model effectively. \revision{To substantiate our claims, we first visualize and compare the learned \TGC and STGCN$(1^{st})$~\cite{yu2018spatio} representations (e.g., $\mathbf{Z}$ mentioned after \shortautoref{eq: fine-grained temporal frequency filtering} in our method) using t-SNE~\cite{van2008visualizing} on two synthetic time series groups with positive and negative correlations.} We create two groups of time series, each with a length of 2000. For the first group, we generate a sinusoidal signal and develop 100 distinct instances with varying amplitudes and injected random noise. Similarly, we generate another group of data based on cosinusoidal oscillation. \shortautoref{subfig: signed relation learning experiments on synthetic data} reveals that STGCN$(1^{st})$ fails to differentiate between the two correlation groups. This is because methods like STGCN$(1^{st})$ aggregate neighborhood information with a single perspective (i.e., low-pass filtration). We further examine the learned embeddings of two groups of randomly sampled time series with opposing trends between \TGC and STGCN$(1^{st})$ on a real-world traffic dataset (\shortautoref{subfig: signed relation learning experiments on pems07 data}), where similar phenomena can be observed.

Additional evaluation can be found in \shortautoref{fig: additional embedding visualization}, where we compare the learned time series embeddings from two \SpecTemGNNs (i.e., Ours and StemGNN~\cite{cao2020spectral}) and \MPSTGNN (i.e., STGCN$(1^{st})$~\cite{yu2018spatio} and Graph WaveNet~\cite{wu2019graph}). It is evident that \SpecTemGNNs excel in learning different signed spatial relations between time series, yielding significantly distinct representations with much higher clustering purities in both cases, further substantiating our claims in the main text.

\begin{figure}[t]
  \centering
  \begin{subfigure}[b]{0.47\linewidth}
  \centering
    \includegraphics[width=\linewidth]{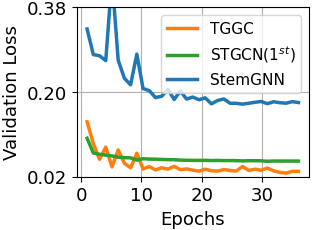}
  \end{subfigure}\hfill
  \begin{subfigure}[b]{0.48\linewidth}
  \centering
    \includegraphics[width=\linewidth]{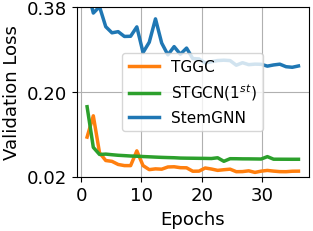}
  \end{subfigure}
  \caption{Model convergence comparison on PeMS07 dataset. \textbf{Left}: $lr=0.01$. \textbf{Right}: $lr=0.001$.}
  \label{fig: model convergencies}
  \vspace{-3mm}
\end{figure}

\subsection{Ablation Studies.}
We perform ablation studies from three perspectives. First, we evaluate the GSFs in \TGC with different polynomial bases (A.1 to A.5) in the first block. Next, we examine other core designs in the second block: B.1 and B.2 apply identical polynomial coefficients in GSFs and trainable weights in temporal FDMs along $D$ dimensions, respectively. B.3 utilizes the same set of TSFs across $N$ variables. B.4 replaces orthogonal space projections with random transformations. B.5 and B.6 separately remove the coarse-grained and fine-grained temporal FDMs. Lastly, we evaluate add-ons that make \TGCfull. C.1 eliminates nonlinearities, and C.2 disables the spectral attention. 

In the first block of results, we validate the discussion in \shortautoref{sec:instant-tgc}: \textbf{(1)} orthogonal polynomials (A.3 to A.5) yield significantly better performance than non-orthogonal alternatives (A.1 and A.2); \textbf{(2)} although the performance gaps between A.3 to A.5 are minor, polynomial bases with orthogonality that hold on more general weight functions tend to result in better performances. The results of B.1 to B.3 support the analysis of multidimensional and multivariate predictions in \shortautoref{subsec:expressive power analysis}, with various degradations observed. In B.4, we see an average \textbf{4\%} MAE and \textbf{7\%} RMSE reduction, confirming the related analysis in \shortautoref{subsec:design of spectral filters}. The results of B.5 and B.6 indicate that both implementations (\shortautoref{eq:coarse-grained-temporal-frequency-filtering} and \shortautoref{eq: fine-grained temporal frequency filtering}) are effective, with fine-grained temporal FDMs icing on the cake. In the last block, we note a maximum \textbf{3.1\%} and \textbf{0.8\%} improvement over \TGC by introducing nonlinearities (C.2) and spectral attention (C.1). Simply combining both leads to even better performance (i.e., \TGCfull).

\begin{figure}[t]
     \centering
     \begin{subfigure}{0.49\linewidth}
         \centering
         \includegraphics[width=\linewidth]{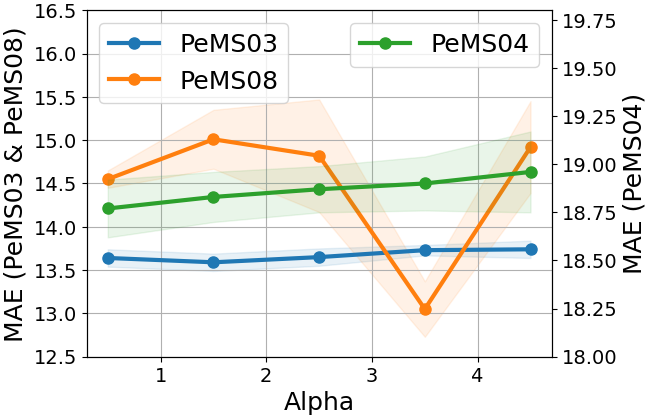}
         \caption{Study on the Gegenbauer parameter $\alpha$.}
         \label{subfig: ps-alpha}
     \end{subfigure}
     \hfill
     \begin{subfigure}{0.46\linewidth}
         \centering
         \includegraphics[width=\linewidth]{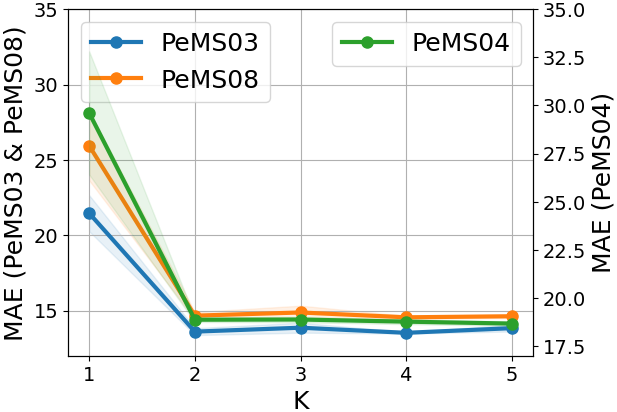}
         \caption{Study on the polynomial degree $K$}
         \label{subfig: ps-k}
     \end{subfigure}
     \vfill
     \begin{subfigure}{0.49\linewidth}
         \centering
         \includegraphics[width=\linewidth]{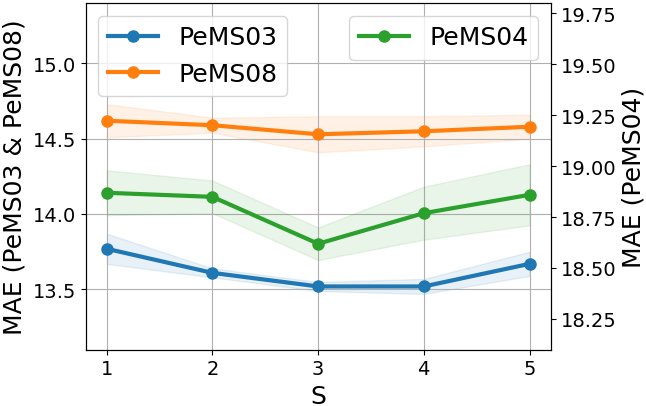}
         \caption{Study on the number of selected modes $S$.}
         \label{subfig: ps-s}
     \end{subfigure}
      \hfill
     \begin{subfigure}{0.49\linewidth}
         \centering
         \includegraphics[width=\linewidth]{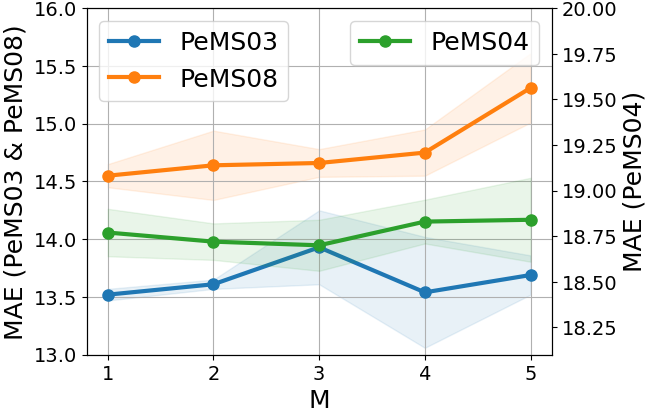}
         \caption{Study on the number of building blocks $M$.}
         \label{subfig: ps-m}
     \end{subfigure}
        \caption{Study on important parameters in \TGC.}
        \label{fig: parameter study}
\end{figure}

\begin{figure}[t]
     \centering
     \begin{subfigure}{0.45\linewidth}
         \centering
         \includegraphics[width=\linewidth]{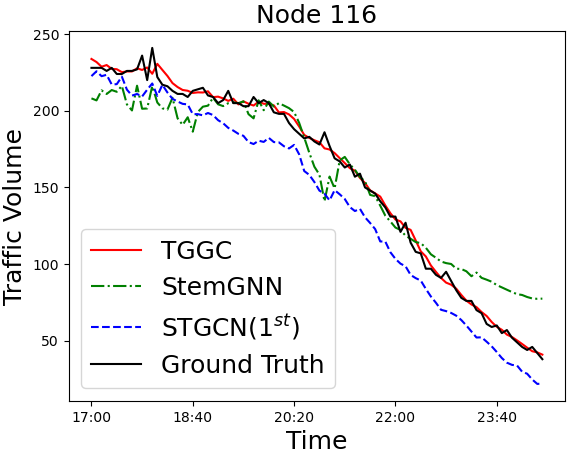}
         \caption{Predictions on 1st selected period on PeMS08}
         \label{subfig: fv-pems08-1-3}
     \end{subfigure}
     \hfill
     \begin{subfigure}{0.45\linewidth}
         \centering
         \includegraphics[width=\linewidth]{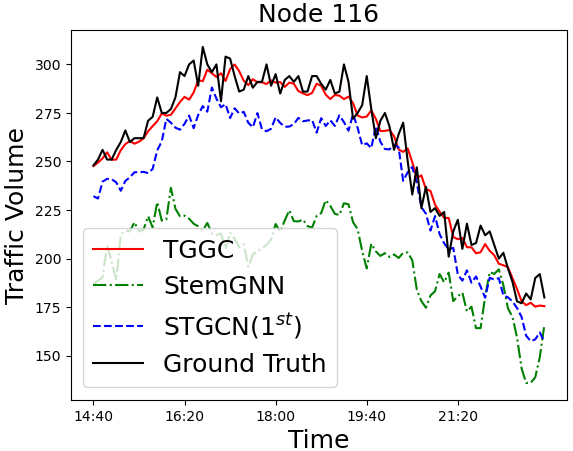}
         \caption{Predictions on 2nd selected period on PeMS08}
         \label{subfig: fv-pems08-2-3}
     \end{subfigure}
     \vfill
     \begin{subfigure}{0.45\linewidth}
         \centering
         \includegraphics[width=\linewidth]{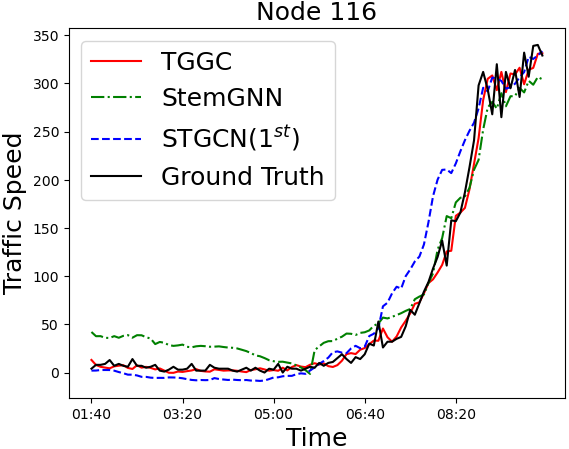}
         \caption{Predictions on 1st selected period on PeMS04}
         \label{subfig: fv-pems04-1-3}
     \end{subfigure}
     \hfill
     \begin{subfigure}{0.45\linewidth}
         \centering
         \includegraphics[width=\linewidth]{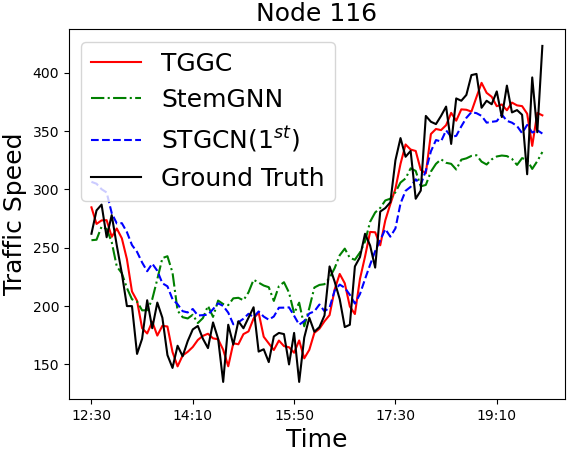}
         \caption{Predictions on 2nd selected period on PeMS04}
         \label{subfig: fv-pems04-2-3}
     \end{subfigure}
        \caption{Showcases of model forecasts on different datasets.}
        \label{fig: forecasting showcases}
\end{figure}

\subsection{Additional Analysis}

\noindent\textbf{Model Convergence.} We compare model convergence between STGCN$(1^{st})$~\cite{yu2018spatio}, StemGNN~\cite{cao2020spectral}, and \TGC across two scenarios with different learning rates in \shortautoref{fig: model convergencies}. Our implementation with Gegenbauer bases has the fastest convergence rate in both cases, further confirming \autoref{theorem: design of spectral graph filters} with ablation studies. Also, as anticipated, STGCN$(1^{st})$ is more tractable than StemGNN w.r.t. the model training due to certain relaxations but at the cost of model effectiveness. \\

\noindent\textbf{Parameter Sensitivity.} The results of parameter study are in \shortautoref{fig: parameter study}, and we have the following observations: \textbf{(1)} adjusting the $\alpha$ in Gegenbauer polynomials impacts model performance variably across datasets. For instance, a smaller $\alpha$ is generally favored in three PeMS datasets, as depicted in \shortautoref{subfig: ps-alpha}; \textbf{(2)} the polynomial degree should not be too small to avoid information loss, as demonstrated by the poor performance when $K=1$ in \shortautoref{subfig: ps-k}. In practice, we set $K$ between 3 and 5, as higher degrees do not bring additional performance gains; \textbf{(3)} similarly, setting $S$ too small or too large is not desirable, as shown in \shortautoref{subfig: ps-s}. This is to prevent information loss and mitigate the impact of noise; \textbf{(4)} stacking more building blocks seems unnecessary; even a single \TGC block already yields competitive performance, according to \shortautoref{subfig: ps-m}. \\

\noindent\textbf{Showcases.} Forecasting visualizations are presented in \shortautoref{fig: forecasting showcases}. We juxtapose \TGC with StemGNN and STGCN$(1^{st})$ using the PeMS08 and PeMS04 datasets. The visualizations capture forecasts from two randomly selected sensors during distinct periods from the test sets. In every instance, \TGC markedly surpasses the other two baselines. \\

\noindent\textbf{Additional Experiments.} Refer to \textbf{\textit{Appendix C}} for details. Specifically, \textbf{(1)} we evaluate \TGC against additional baselines on other time series benchmarks; \textbf{(2)} we provide the statistics of our forecasting results.
\section{Conclusion}\label{sec:conclusion}
In this study, we provide the general formulation of spectral-temporal graph neural networks (\SpecTemGNNs), laying down a theoretical framework for this category of methods. The key insights derived include: \textbf{(1)} under modest assumptions, \SpecTemGNNs can achieve universality; \textbf{(2)} the use of orthogonal bases and individual spectral filters are pivotal in crafting potent GNN-based time series models. To validate our theoretical findings, we introduce an innovative yet straightforward spectral-temporal graph neural network (\TGC) and its enhanced variant (\TGCfull). Both consistently surpass the majority of baseline methods across diverse real-world benchmarks. Our findings shed light on the follow-up research and pave the way for devising a broader array of provably expressive \SpecTemGNNs. \revisiontwo{A limitation of our approach is that \SpecTemGNNs may face challenges in handling specific types of time series data, such as the underlying graph topology is highly symmetric and node attributes exhibit strong non-stationary or non-periodic behavior. These cases, while rare, can pose difficulties for our method and will necessitate further refinement of the development of more adaptive models. In addition to addressing this limitation, future research will also explore} specific scenarios such as time-evolving graph structures and investigating the applicability of our theories to other tasks, such as time series classification and anomaly detection.

\section*{Acknowledgments}
This research was partly funded by Australian Research Council (ARC) under grants FT210100097
and DP240101547 and the CSIRO – National Science Foundation (US) AI Research Collaboration
Program.

\bibliographystyle{IEEEtran}
\bibliography{IEEEabrv,reference}

\clearpage
\newpage
\setcounter{page}{1}

\appendices
\section{Proofs}\label{appx: proofs}

\subsection{Proof of \autoref{theorem: linear stgnn}}
\textbf{\autoref{theorem: linear stgnn}}.
\textit{A linear \SpecTemGNN can produce arbitrary dimensional time series representations at any valid time iff: \textbf{(1)} $\Hat{\mathbf{L}}$ has no repeated eigenvalues; \textbf{(2)} $\mathbf{X}$ encompasses all frequency components with respect to the graph spectrum; \textbf{(3)} $\mathcal{T}_\phi(\cdot)$ can express any single-dimensional univariate time series.}

\begin{proof}
Let us assume that $t=1$ (i.e., there is only one snapshot $\mathcal{G}_t$ of the graph $\{\mathcal{G}_t\}_{t=0}^{T-1}$), $\mathcal{T}_\phi(\cdot)$ reduces to linear functions of a graph snapshot and the linear spectral-temporal GNN reduces to a linear spectral GNN, which can be equivalently written as: 

\begin{equation}
    \mathbf{Z}_t = g_\theta(\Hat{\mathbf{L}}) \mathbf{X}_t \mathbf{\Phi} \in \mathbb{R}^{N \times D}.
\end{equation}

We first prove the universality theorem of linear spectral GNNs in a graph snapshot on the basis of Theorem 4.1 in \cite{wang22jacobi}. In other words, assuming $\Tilde{\mathbf{X}}_t = \mathbf{U}^\top \mathbf{X}_t$ has non-zero row vectors and $\Hat{\mathbf{L}}$ has unique eigenvalues, we first aim to prove that for any $\mathbf{Z}_{t,d} \in \mathbb{R}^{N \times 1}$, there is a linear spectral GNN to produce it. 
We assume there exists $\boldsymbol{\phi}^* \in \mathbb{R}^D$ s.t. all elements in $\Tilde{\mathbf{X}}_t\boldsymbol{\phi}^*$ are non-zero. Considering a case where $(\Tilde{\mathbf{X}}_t\boldsymbol{\phi})_i = 0$ and letting the solution space to be $\mathbb{S}_i$, we know that $\mathbb{S}_i$ is a proper subspace of $\mathbb{R}^D$ as the $i$-th row of $\Tilde{\mathbf{X}}_t$ is non-zero. Therefore, $\mathbb{R}^D \setminus \cup_{i=1}^{N}\mathbb{S}_i \neq \emptyset$, and we know that all vectors $\boldsymbol{\phi}$ in $\mathbb{R}^D \setminus \cup_{i=1}^{N}\mathbb{S}_i$ are valid to form $\boldsymbol{\phi}^*$.
We then filter $\Tilde{\mathbf{X}}_t\boldsymbol{\phi}^*$ to get $\mathbf{Z}_{t,d}$. Firstly, we let $\Tilde{\mathbf{Z}}_{t,d} = \mathbf{U}^\top \mathbf{Z}_{t,d}$ and assume there is a polynomial with $N-1$ order:

\begin{equation}
\begin{aligned}
    p_i &:= g_\theta(\lambda_i), \\
    &= \sum_{k=0}^{N-1} \theta_k \lambda_i^k \ \ \ \text{s.t.}\ \ \  p_i = \Tilde{\mathbf{Z}}_{t,d}[i] / (\Tilde{\mathbf{X}}_t \boldsymbol{\phi}^*)_i \ \ ,\  \forall i \in \{1, \cdots, N\}. 
\end{aligned}
\end{equation}

On this basis, the polynomial coefficients $\boldsymbol{\theta}$ is the solution of a linear system $\mathbf{B}\boldsymbol{\theta}=\mathbf{p}$ where $\mathbf{B}_{i,j} = \lambda_i^{j-1}$. Since $\lambda_i$ are different from each other, $\mathbf{B}^\top$ turns to a nonsingular Vandermonde matrix, where a solution $\boldsymbol{\theta}$ always exists. Therefore, a linear spectral GNN can produce any one-dimensional prediction under certain assumptions.

The above proof states that linear spectral GNNs can produce any one-dimensional prediction if $\Hat{\mathbf{L}}$ has no repeated eigenvalues (i.e., condition 1) and the node features $\mathbf{X}$ contain all frequency components w.r.t. graph spectrum (i.e., condition 2). 
When $t \geq 2$, $\mathcal{T}_\phi(\cdot)$ turns to linear functions over all historical observations of graph snapshots. In order to distinguish between different historical graph snapshots, $\mathcal{T}_\phi(\cdot)$ must be universal approximations of all historical graph snapshots, implying that $\mathcal{T}_\phi(\cdot)$ is able to express any one-dimensional univariate time series (i.e., condition 3).
\end{proof}

\subsection{Proof of \autoref{theorem: linear stgnn and wl-test}}
\textbf{\autoref{theorem: linear stgnn and wl-test}}.
\textit{For a linear \SpecTemGNN \revision{with valid temporal FDMs that can express arbitrary 1-dimensional univariate time series} and a $K$-degree polynomial basis in its GSFs, $\forall u,v \in \mathbb{V}, \mathbf{Z}_t[u] = \mathbf{Z}_t[v]$ if $\mathbf{C}^{(K+1)}(u,t) = \mathbf{C}^{(K+1)}(v,t)$. $\mathbf{Z}_t[i]$ and $\mathbf{C}^{(K)}(i,t)$ represent node $i$'s embedding at time $t$ in such a GNN and the $K$-step temporal 1-WL test, respectively.}

\begin{proof}
Given valid temporal frequency-domain models (i.e., space projectors and TSFs) and a $K$-degree polynomial filter function, the prediction of a linear \SpecTemGNN can be formulated as follows.

\begin{equation}
    \mathbf{Z} = \mathcal{T}_{\phi}\left(\sum_{k=0}^K \boldsymbol{\Theta}_{k} P_k(\hat{\mathbf{L}}) \mathbf{X}\right).
\label{eq:zeta}
\end{equation}

For ease of reading, we redefine $\mathcal{T}_{\phi}(\cdot)$ as the combination of space projections and TSFs in the following proof. 
Let us assume $t=1$ (i.e., there is only one snapshot $\mathcal{G}_t$ of the graph $\{\mathcal{G}_t\}_{t=0}^{T-1}$), $\mathcal{T}_\phi(\cdot)$ reduces to linear functions of a single graph snapshot and a linear \SpecTemGNN reduces to a linear spectral GNN. Using the framework in \cite{xu2019graph}, \shortautoref{eq:zeta} can be viewed as a $k+1$-layer GNN. 
The output of the last layer in GNN produces the output of linear spectral GNNs \cite{wang22jacobi}. According to the proof of Lemma 2 in \cite{xu2019graph}, if WL node labels $\mathbf{C}^{(K+1)}(u)=\mathbf{C}^{(K+1)}(v)$, the corresponding GNN's node features should be the same at any iteration. Therefore, for all nodes $\forall u,v \in \mathbb{V}, \mathbf{Z}[u] = \mathbf{Z}[v]$ if $\mathbf{C}^{(K+1)}(u) = \mathbf{C}^{(K+1)}(v)$.

When $t \geq 2$, we have a DTDG defined as a sequence of graph snapshots $\left(\mathcal{G}_1, \mathcal{G}_2, \ldots\right)$ that are sampled at regular intervals, and each snapshot is a static graph. Note that any DTDGs can be equivalently converted to continuous-time temporal graphs (CTDGs). The CTDG can be equivalently viewed as time-stamped multi-graphs with timestamped edges, i.e., $\mathcal{G}(t)=$ $\left\{\left(u_k, v_k, t_k\right) \mid t_k<t\right\}$. 
According to Proposition 6 in \cite{souza2022provably}, the expressive power of dynamic GNN with injective message passing is bounded by the temporal WL test on $\mathcal{G}(t)$. Since $\mathcal{T}_\phi(\cdot)$ is a set of linear functions over all historical observations of graph snapshots, i.e., $\mathcal{T}_\phi(\cdot)$ represents linear transformations of $\mathcal{G}(t)$. 
The defined \SpecTemGNN, i.e., $\mathbf{Z} = \mathcal{T}_{\phi}(\sum_{k=0}^K \boldsymbol{\Theta}_{k} P_k(\hat{\mathbf{L}}) \mathbf{X})$, should be as expressive as dynamic GNN with injective message passing. Hence, if temporal WL node labels $\mathbf{C}^{(K+1)}(u,t)=\mathbf{C}^{(K+1)}(v,t)$, the corresponding GNN's node features should be the same at any timestamp $t$ and at any iteration. Therefore, for all nodes $\forall u,v \in \mathbb{V}, \mathbf{Z}_t[u] = \mathbf{Z}_t[v]$ if $\mathbf{C}^{(K+1)}(u,t) = \mathbf{C}^{(K+1)}(v,t)$. 
\end{proof}

\subsection{Proof of \autoref{prop: narrow temporal 1-wl down}}
\textbf{\autoref{prop: narrow temporal 1-wl down}}.
\textit{If a discrete-time dynamic graph with a fixed graph topology at time $t$ has no repeated eigenvalues in its normalized graph Laplacian and has no missing frequency components in each snapshot, then the temporal 1-WL is able to differentiate all non-isomorphic nodes at time $t$.}

\begin{proof}
Assume there are no repeated eigenvalues and missing frequency components w.r.t. graph spectrum in a DTDG with fixed topology and time-evolving features, i.e., $\{\mathcal{G}_t\}_{t=0}^{T-1}$.

According to Corollary 4.4 in \cite{wang22jacobi}, we know that if a graph has no repeated eigenvalues and missing frequency components, then 1-WL test can differentiate any pair of non-isomorphic nodes. We denote the colors of two nodes $u$ and $v$ in $\mathcal{G}_t$ after $L$ 1-WL interactions as $\mathbf{C}^{(L)}(u,t)$ and $\mathbf{C}^{(L)}(v,t)$ s.t. $\mathbf{C}^{(L)}(u,t) \neq \mathbf{C}^{(L)}(v,t)$ if $u$ and $v$ are non-isomorphic.
On this basis, we consider two scenarios in $\mathcal{G}_{t+1}$: \textbf{(1)} two or more non-isomorphic nodes have identical initial colors; \textbf{(2)} none of the non-isomorphic nodes have identical colors. Under the assumptions in this proposition, the 1-WL test can differentiate $u$ and $v$ in $\mathcal{G}_{t+1}$ on both cases with different \\

\noindent $\mathbf{C}^{(L)}(u,t+1) := \textsc{Hash}(c^{(L-1)}(u,t+1), \{\!\!\{ c^{(L-1)}(m,t+1) : e_{u,m,t+1} \in \mathbb{E}(\mathcal{G}_{t+1}) \}\!\!\})$ \\

and \\

\noindent $\mathbf{C}^{(L)}(v,t+1) := \textsc{Hash}(c^{(L-1)}(v,t+1), \{\!\!\{ c^{(L-1)}(m,t+1) : e_{v,m,t+1} \in \mathbb{E}(\mathcal{G}_{t+1}) \}\!\!\})$. \\

Therefore, no matter whether $\mathbf{C}^{(L)}(u,t)$ and $\mathbf{C}^{(L)}(v,t)$ are identical or not (they are different in fact as mentioned), we have nonidentical \\

\noindent $\mathbf{C}^{(L)}(u,t+1) := \textsc{Hash}(c^{(L-1)}(u,t+1), c^{(L-1)}(u,t), \{\!\!\{ c^{(L-1)}(m,t+1) : e_{u,m,t+1} \in \mathbb{E}(\mathcal{G}_{t+1}) \}\!\!\})$ \\

and \\

\noindent $\mathbf{C}^{(L)}(v,t+1) := \textsc{Hash}(c^{(L-1)}(v,t+1), c^{(L-1)}(v,t), \{\!\!\{ c^{(L-1)}(m,t+1) : e_{v,m,t+1} \in \mathbb{E}(\mathcal{G}_{t+1}) \}\!\!\})$ \\

\noindent in the temporal 1-WL test, where $\mathbf{C}^{(L)}(u,t) := c^{(L-1)}(u,t)$ and $\mathbf{C}^{(L)}(v,t) := c^{(L-1)}(v,t)$. 
\end{proof}

\subsection{Proof of \autoref{prop: no graph automorphism}}
\textbf{\autoref{prop: no graph automorphism}}.
\textit{If a discrete-time dynamic graph with a fixed graph topology has no multiple eigenvalues in its normalized graph Laplacian and has no missing frequency components in each snapshot, then no automorphism exists.}

\begin{proof}
    Given a DTDG $\{\mathcal{G}_t\}_{t=0}^{T-1}$ consists of $T$ static graph snapshots with fixed graph topology and time-evolving node features, we first prove that all pairs of nodes are non-isomorphic. 
    
    In a snapshot $\mathcal{G}_t$, assume there is a permutation matrix $\mathbf{P}$, we have
    \begin{equation}
        \Hat{\mathbf{L}} := \mathbf{P}^\top \Hat{\mathbf{L}} \mathbf{P} = \mathbf{P} \mathbf{U} \mathbf{\Lambda} \mathbf{U}^\top \mathbf{P}^\top,
    \end{equation}
    and we know
    \begin{equation}
        \mathbf{\Lambda} := \mathbf{U}^\top \mathbf{P} \mathbf{U} \mathbf{\Lambda} \mathbf{U}^\top \mathbf{P}^\top \mathbf{U} = \mathbf{V} \mathbf{\Lambda} \mathbf{V}^\top,
    \end{equation}
    where $\mathbf{V}$ is an orthogonal matrix. Since all diagonal elements in $\mathbf{\Lambda}$ are different because we assume no repeated eigenvalues, then the eigenspace of each eigenvalue has only one dimension \cite{wang22jacobi}; thus, we have $\mathbf{U}^\top \mathbf{P} \mathbf{U} = \mathbf{D}$, where $\mathbf{D}$ is a diagonal matrix s.t. $\mathbf{V} := \mathbf{D}$ with $\pm 1$ elements.  
    Now considering the node features $\mathbf{X}_t := \mathbf{P}\mathbf{X}_t$, we have $\Hat{\mathbf{X}}_t = \mathbf{V}\Hat{\mathbf{X}}_t$; thus $(\mathbf{I}-\mathbf{D})\Hat{\mathbf{X}}_t = 0$ based on the above discussion. If there are no missing frequency components in $\Hat{\mathbf{X}}_t$, i.e., no zero row vectors, we have $\mathbf{D} = \mathbf{I}$ and know that
    \begin{equation}
        \mathbf{P} = \mathbf{U} \mathbf{D} \mathbf{U}^\top = \mathbf{I}.
    \end{equation}
    
    Hence, we prove that all nodes in a graph snapshot $\mathcal{G}_t$ are non-isomorphic.
    In $\{\mathcal{G}_t\}_{t=0}^{T-1}$, we have $\mathbf{V} := \mathbf{D}$ always holds across all snapshots if its normalized graph Laplacian has no repeated eigenvalues. On this basis, if there are no missing frequency components by giving $\mathbf{X}_t, \forall t \in \{0, 1, \ldots, T-1\}$, all pairs of nodes are non-isomorphic in an attributed DTDG with fixed graph topology.

\end{proof}

\subsection{Proof of \autoref{theorem: design of spectral graph filters}}
\textbf{\autoref{theorem: design of spectral graph filters}}.
\textit{
For a linear \SpecTemGNN optimized with mean squared loss, any complete polynomial bases result in the same expressive power, but an orthonormal basis guarantees the maximum convergence rate if its weight function matches the graph signal density.
}

\begin{proof}
Directly analyzing \shortautoref{eq: general formula} is complex and unnecessary to study the effectiveness of different polynomial bases when learning time series relations at each time step. Since optimizing the spectral-temporal GNNs formulated in \shortautoref{eq: general formula} can be understood as a two-step (i.e., graph-then-temporal) optimization problem, we directly analyze the optimization of $\mathbf{\Theta}$ w.r.t. the formulation below based on the squared loss $\mathcal{L}=\frac{1}{2} || \mathbf{Z}_t - \mathbf{Y}_t ||^{2}_{F}$ on a graph snapshot $\mathcal{G}_t$ with the target $\mathbf{Y}_t$.

\begin{equation}
    \mathbf{Z}_t = \sum_{k=0}^{K} \mathbf{\Theta}_k P_k(\Hat{\mathbf{L}}) \mathbf{X}_t.
\end{equation}

This is a convex optimization problem, thus the convergence rate of gradient descent relates to the condition number of the Hessian matrix \cite{boyd2004convex}. In other words, the convergence rate reaches the maximum if $\kappa{(\mathbf{H})}$ reaches the minimum. We have $\mathbf{H}_{k_1,k_2}$ defined as follows that is similar in \cite{wang22jacobi}.

\begin{equation}
\begin{aligned}
    \frac{\partial \mathcal{L}}{\partial \mathbf{\Theta}_{k_1} \partial \mathbf{\Theta}_{k_2}} &= \mathbf{X}_{t}^\top P_{k_2}(\Hat{\mathbf{L}}) P_{k_1}(\Hat{\mathbf{L}}) \mathbf{X}_{t}, \\
    &= \sum_{i=1}^{n} P_{k_2}(\lambda_i) P_{k_1}(\lambda_i) \Tilde{\mathbf{X}}_{t}[\lambda_i].
\end{aligned}
\end{equation}

This equation can be written as a Riemann sum as follows.

\begin{equation}
    \frac{\partial \mathcal{L}}{\partial \mathbf{\Theta}_{k_1} \partial \mathbf{\Theta}_{k_2}} = \sum_{i=1}^{n} P_{k_2}(\lambda_i) P_{k_1}(\lambda_i) \frac{F(\lambda_i) - F(\lambda_{i-1})}{\lambda_i - \lambda_{i-1}} (\lambda_i - \lambda_{i-1}).
\end{equation}

In the above formula, $F(\lambda_i) := \sum_{\lambda_j \leq \lambda_i} (\Tilde{\mathbf{X}}_{t}[\lambda_j])^2$ and $\frac{F(\lambda_i) - F(\lambda_{i-1})}{\lambda_i - \lambda_{i-1}}$ denotes the graph signal density at the frequency $\lambda_i$. When $n \rightarrow \infty$, we have the $(k_1, k_2)$ element in $\mathbf{H}$ rewrite as follows.

\begin{equation}
    \mathbf{H}_{k_1,k_2} =  \int_{\lambda=0}^{2} P_{k_2}(\lambda) P_{k_1}(\lambda) \frac{\Delta F(\lambda)}{\Delta \lambda} \,d\lambda.
\end{equation}

We know that $\kappa{(\mathbf{H})}$ reaches the minimum if $\mathbf{H}$ is a diagonal matrix, which tells that the polynomial bases, e.g., $P_{k_1}(\lambda)$ and $P_{k_2}(\lambda)$, should be orthogonal w.r.t. the weight function $\frac{\Delta F(\lambda)}{\Delta \lambda}$.

\end{proof}

\subsection{Proof of \autoref{lemma: linear filtering operator new}}
\textbf{\autoref{lemma: linear filtering operator new}}.
\textit{Suppose the projection of $\mathbf{A}$ by $\mathbf{A}'$ is $P_{\mathbf{A}'}(\mathbf{A})$, and the coherence measure of $\mathbf{A}$ is $\mu(\mathbf{A})=\Omega(k/N)$, then with a high probability, the error between $\mathbf{A}\mathbf{W}$ and $P_{\mathbf{A}'}(\mathbf{A})\mathbf{W}$ is bounded by $||\mathbf{A}\mathbf{W} - P_{\mathbf{A}'}(\mathbf{A})\mathbf{W}||_F \leq (1 + \epsilon)||\mathbf{W}||_F||\mathbf{A} - \mathbf{A}_k||_F$ if $S=O(k^2/\epsilon^2)$.}

\begin{proof}
Similar to the analysis in Theorem 3 from \cite{drineas2008relative} and Theorem 1 from \cite{zhou22fedformer}, we have

\begin{equation}\label{eq_bound1}
\begin{aligned}
||\mathbf{A}\mathbf{W} - P_{\mathbf{A}'}(\mathbf{A})\mathbf{W}||_F 
& \leq ||\mathbf{W}||_F ||\mathbf{A} - P_{\mathbf{A}'}(\mathbf{A})||_F \\
& =    ||\mathbf{W}||_F ||\mathbf{A} - \mathbf{A}'(\mathbf{A}')^{\dagger}\mathbf{A}||_F \\
& \leq ||\mathbf{W}||_F ||\mathbf{A} - (\mathbf{A}\mathbf{S}^{\top})(\mathbf{A}_k\mathbf{S}^{\top})^{\dagger}\mathbf{A}_k||_F. 
\end{aligned} 
\end{equation}

Then, following Theorem 5 from \cite{drineas2008relative}, if $S = O(k^2/\epsilon^2\times \mu(\mathbf{A})N/k)$, we can obtain the following result with a probability at least $0.7$

\begin{equation}\label{eq_bound2}
||\mathbf{A} - (\mathbf{A}\mathbf{S}^{\top})(\mathbf{A}_k\mathbf{S}^{\top})^{\dagger}\mathbf{A}_k||_F \leq (1 + \epsilon)||\mathbf{A} - \mathbf{A}_k||_F.
\end{equation}

Since $\mu(\mathbf{A})=\Omega(k/N)$, when $S=O(k^2/\epsilon^2)$ together with \shortautoref{eq_bound1} and \shortautoref{eq_bound2},  we can obtain the final bound as 

\begin{equation}
||\mathbf{A}\mathbf{W} - P_{\mathbf{A}'}(\mathbf{A})\mathbf{W}||_F \leq (1 + \epsilon)||\mathbf{W}||_F||\mathbf{A} - \mathbf{A}_k||_F.
\end{equation}
\end{proof}
\section{Additional Experimental Settings}\label{appx:additional-experiment-setting}

The detailed hyperparameter configurations are in \shortautoref{tab: short-term forecasting hyperparameters} and \shortautoref{tab: long-term forecasting hyperparameters}.

\begin{table}[h]
\centering
\caption{The hyperparameter setting of our method for short-term time series forecasting.}
\scalebox{0.65}{
\begin{tabular}{l|ccccccc}
\toprule
Hyperparameters          & PeMS03               & PeMS04               & PeMS07               & PeMS08               & Electricity          & Solar                & ECG                  \\ \hline
Gegenbauer parameter $\alpha$             &3.08                      &0.47                      &1                      &1                      &1.2                      &1.2                      &1.2                      \\
\# polynomial degree $K$    &4                      &4                      &4                      &4                      &4                      &4                      &4                      \\
\# selected component $S$ &5                      &4                      &5                      &5                      &5                      &5                      &5                      \\
\# building block $M$              &2                      &2                      &2                      &2                      &2                      &2                      &2                      \\

Training batch size B              &32      &64     &50    &50    &50    &50    &50 \\
Model learning rate $\eta$            &0.0003   &0.001  &0.001  &0.001  &0.001  &0.001  &0.001 \\ \bottomrule
\end{tabular}}
\label{tab: short-term forecasting hyperparameters}
\end{table}

\begin{table}[h]
\centering
\caption{The hyperparameter setting of our method for long-term time series forecasting.}
\scalebox{0.75}{
\begin{tabular}{l| p{25 pt} p{25 pt} p{25 pt} p{25 pt} p{25 pt} p{25 pt} p{25 pt}}
\toprule
Hyperparameters & $\alpha$ & $\beta$ & $K$ & $S$ & $M$ & $B$ & $\eta$ \\ \hline
Electricity     &1       &1      &4                       &14                          &5                      &50                       &0.001                     \\
Solar-Energy    &1       &1      &4                       &50                          &5                      &50                       &0.001                     \\
Weather         &0.81       &0.90      &4                       &26                          &3                      &64                       &0.0003                     \\ \bottomrule
\end{tabular}}
\label{tab: long-term forecasting hyperparameters}
\end{table}
\section{Additional Results}\label{appx:additional-result}

\subsection{Additional Forecasting Results}

\begin{table}[t]
\centering
\setlength\extrarowheight{2pt}
\caption{Additional short-term forecasting results on three time series benchmarks, where we follow \cite{cao2020spectral} for the experimental setting and baseline results.
We use the \textcolor{red}{\textbf{bold}} and \textcolor{blue}{\underline{underline}} fonts to indicate the best and second-best results.}
\scalebox{0.9}{
\begin{tabular}{p{50pt} *{6}{p{0.73cm}}}
\toprule
\multicolumn{1}{c}{\multirow{2}{*}{Method}} & MAE         & RMSE         & MAE         & RMSE         & MAE         & RMSE         \\ \cline{2-3} \cline{4-5} \cline{6-7}
\multicolumn{1}{c}{}                        & \multicolumn{2}{c}{\textit{Electricity}} & \multicolumn{2}{c}{\textit{Solar}} & \multicolumn{2}{c}{\textit{ECG}} \\ \hline
FC-LSTM~\cite{sutskever2014sequence}                                     &0.62             &0.20              &0.13             &0.19              &0.32             &0.54  \\
TCN~\cite{bai2018empirical}                                         &0.07             &0.51              &0.06             &0.06              &0.10             &0.30   \\
LSTNet~\cite{lai2018modeling}                                      &0.06             &0.07              &0.07             &0.19              &0.08             &0.12  \\
DeepState~\cite{rangapuram2018deep}                                   &0.06             &0.67              &0.06             &0.25              &0.09            &0.76  \\
DeepGLO~\cite{sen2019think}                                     &0.08             &0.14              &0.09             &0.14              &0.09             &0.15  \\
SFM~\cite{zhang2017stock}                                      &0.08             &0.13              &0.05             &0.09             &0.17             &0.58  \\
GWNet~\cite{wu2019graph}                               &0.07             &0.53              &0.09             &0.14              &0.09             &0.15   \\
StemGNN~\cite{cao2020spectral}                                     &\textcolor{red}{\textbf{0.04}}             &\textcolor{red}{\textbf{0.06}}              &\textcolor{blue}{\underline{0.03}}             &\textcolor{blue}{\underline{0.07}}              &\textcolor{blue}{\underline{0.05}}             &\textcolor{blue}{\underline{0.07}}  \\
\TGC \textbf{(Ours)}                  &\textcolor{blue}{\underline{0.05}}            &\textcolor{blue}{\underline{0.07}}              &\textcolor{red}{\textbf{0.02}}             &\textcolor{red}{\textbf{0.04}}              &\textcolor{red}{\textbf{0.04}}             &\textcolor{red}{\textbf{0.06}}  \\                
\bottomrule
\end{tabular}}
\label{tab: additional-short-term-results}
\end{table}

We provide our supplementary short-term forecasting results in \shortautoref{tab: additional-short-term-results}, from which the following observations can be made: \textbf{(1)} \TGC, in most cases, outperforms all baseline methods by substantial margins, with an average improvement of 35\% compared to the best deep time series baselines; \textbf{(2)} it generally outperforms StemGNN, although the performance gaps are not very significant under this experimental setting.

\subsection{Main Result Statistics}

\shortautoref{tab: short-term-statistics} and \shortautoref{tab: long-term-statistics} present the average performances and 95\% confidence intervals for our results reported in \shortautoref{tab:short-term-part1}, \shortautoref{tab:short-term-part2}, and \shortautoref{tab: long-term forecasting} with 5 individual runs.

\begin{table}[h]
\small
\centering
\setlength\extrarowheight{2pt}
\caption{Short-term forecasting results showing our average performances $\pm 95\%$ confidence intervals, respectively.}
\scalebox{0.7}{
\begin{tabular}{cc|cccc}
\toprule
\multicolumn{2}{c}{\multirow{2}{*}{Method}}                       & \multicolumn{2}{c}{\TGC} & \multicolumn{2}{c}{\TGCfull} \\ \cline{3-6}
\multicolumn{2}{c}{}                       & MAE        & RMSE       & MAE        & RMSE        \\ \midrule
\multicolumn{1}{c|}{\multirow{2}{*}{\rotatebox{0}{\textit{PeMS03}}}} & 3  & 13.52 $\pm$ 0.226          & 21.74 $\pm$ 0.687          & -          & -           \\
\multicolumn{1}{c|}{}                        & 12 & -          & -          & 16.22 $\pm$ 0.475          & 27.07 $\pm$ 0.623           \\ \hline
\multicolumn{1}{c|}{\multirow{2}{*}{\rotatebox{0}{\textit{PeMS04}}}} & 3  & 18.77 $\pm$ 0.142          & 29.92 $\pm$ 0.150          & -          & -           \\
\multicolumn{1}{c|}{}                        & 12 & -          & -          & 20.00 $\pm$ 0.300          & 32.10 $\pm$ 0.452           \\ \hline
\multicolumn{1}{c|}{\multirow{2}{*}{\rotatebox{0}{\textit{PeMS07}}}} & 3  & 1.92 $\pm$ 0.029          & 3.35 $\pm$ 0.093          & -          & -           \\
\multicolumn{1}{c|}{}                        & 12 & -          & -          & 2.81 $\pm$ 0.035          & 5.58 $\pm$ 0.066           \\ \hline
\multicolumn{1}{c|}{\multirow{2}{*}{\rotatebox{0}{\textit{PeMS08}}}} & 3  & 14.55 $\pm$ 0.245          & 22.73 $\pm$ 0.274          & -          & -           \\
\multicolumn{1}{c|}{}                        & 12 & -          & -          & 16.54 $\pm$ 0.981          & 26.10 $\pm$ 0.825           \\ \bottomrule
\end{tabular}
}
\label{tab: short-term-statistics}
\end{table}

\begin{table}[h]
\small
\centering
\setlength\extrarowheight{2pt}
\caption{Long-term forecasting results showing our average performances $\pm 95\%$ confidence intervals, respectively.}
\scalebox{0.7}{
\begin{tabular}{cc|cc}
\toprule
\multicolumn{2}{c}{\multirow{2}{*}{Method}}            & \multicolumn{2}{c}{\TGCfull} \\ \cline{3-4} 
\multicolumn{2}{c}{}                                   & MAE        & RMSE        \\ \midrule
\multicolumn{1}{c|}{\multirow{3}{*}{\textit{Electricity}}} & 96  &   0.293 $\pm$ 0.004         &  0.425 $\pm$ 0.006           \\
\multicolumn{1}{c|}{}                             & 192 &  0.303 $\pm$ 0.006           &   0.440 $\pm$ 0.007          \\
\multicolumn{1}{c|}{}                             & 336 &  0.313 $\pm$ 0.007          &   0.470 $\pm$ 0.005          \\ \hline
\multicolumn{1}{c|}{\multirow{3}{*}{\textit{Weather}}}     & 96  &  0.235 $\pm$ 0.014          &   0.408 $\pm$ 0.029          \\
\multicolumn{1}{c|}{}                             & 192 &  0.286 $\pm$ 0.021          &   0.468 $\pm$ 0.041          \\
\multicolumn{1}{c|}{}                             & 336 &  0.317 $\pm$ 0.019          &   0.515 $\pm$ 0.036          \\ \hline
\multicolumn{1}{c|}{\multirow{3}{*}{\textit{Solar}}}       & 96  &  0.242 $\pm$ 0.005          &   0.443 $\pm$ 0.009          \\
\multicolumn{1}{c|}{}                             & 192 &  0.263 $\pm$ 0.004          &   0.470 $\pm$ 0.010          \\
\multicolumn{1}{c|}{}                             & 336 &  0.271 $\pm$ 0.004          &   0.478 $\pm$ 0.009          \\ \bottomrule
\end{tabular}
}
\label{tab: long-term-statistics}
\end{table}

\end{document}